\documentclass[lettersize,journal]{IEEEtran}
\usepackage{amsmath,amsfonts}
\usepackage{algorithmic}
\usepackage{algorithm}
\usepackage{array}
\usepackage[caption=false,font=normalsize,labelfont=sf,textfont=sf]{subfig}
\usepackage{textcomp}
\usepackage{stfloats}
\usepackage{url}
\usepackage{verbatim}
\usepackage{graphicx}
\usepackage{cite}
\usepackage{multirow}
\usepackage{amsmath}
\usepackage{amssymb}
\usepackage[table]{xcolor}
\usepackage{xcolor}

\usepackage{capt-of}
\DeclareMathOperator*{\argmax}{arg\,max}

\makeatletter
\def\endthebibliography{%
  \def\@noitemerr{\@latex@warning{Empty `thebibliography' environment}}%
  \endlist
}
\makeatother

\begin{document}

\title{Inverse Image Frequency for Long-tailed Image Recognition}

\author{Konstantinos Panagiotis Alexandridis$^{1,2}$, Shan Luo$^{1,2,*}$, Anh Nguyen$^{2}$, Jiankang Deng$^{3}$ and Stefanos Zafeiriou$^{3}$

\thanks{Manuscript received: 5th September, 2022.}

\thanks{$^{1}$K. P. Alexandridis and S. Luo are with Department of Engineering, King's College London, London WC2R 2LS, United Kingdom. E-mails: \tt\small\{konstantinos.alexandridis, shan.luo\}@kcl.ac.uk.}%

\thanks{$^{2}$K. P. Alexandridis, A. Nguyen and S. Luo are with the Department of Computer Science, University of Liverpool, Liverpool L69 3BX, United Kingdom. E-mails: \tt\small\{konsa15,anh.nguyen, shan.luo\}@liverpool.ac.uk.}

\thanks{$^{3}$J. Deng and S. Zafeiriou are with Department of Computing, Imperial College London, London SW7 2AZ, United Kingdom. E-mails: \tt\small\{j.deng16, s.zafeiriou\}@ic.ac.uk.}%
\thanks{$^{*}$Corresponding author.}%

\thanks{This paper has supplementary downloadable material available at http://ieeexplore.ieee.org., provided by the author. The material includes a pdf file with the Appendix. Contact shan.luo@kcl.ac.uk for further questions about this work.}
}

\markboth{IEEE Transactions on Image Processing,~Vol.~xx, No.~xx, September~2022}%
{Shell \MakeLowercase{\textit{et al.}}: A Sample Article Using IEEEtran.cls for IEEE Journals}



\maketitle

\begin{abstract}
The long-tailed distribution is a common phenomenon in the real world. Extracted large scale image datasets inevitably demonstrate the long-tailed property and models trained with imbalanced data can obtain high performance for the over-represented categories, but struggle for the under-represented categories, leading to biased predictions and performance degradation. To address this challenge, we propose a novel de-biasing method named \textit{Inverse Image Frequency (IIF)}. IIF is a multiplicative margin adjustment transformation of the logits in the classification layer of a convolutional neural network. Our method achieves stronger performance than similar works and it is especially useful for downstream tasks such as long-tailed instance segmentation as it produces fewer false positive detections. Our extensive experiments show that IIF surpasses the state of the art on many long-tailed benchmarks such as ImageNet-LT, CIFAR-LT, Places-LT and LVIS, reaching $55.8\%$ top-1 accuracy with ResNet50 on ImageNet-LT and $26.3\%$ segmentation AP with MaskRCNN ResNet50 on LVIS. Code available at \url{https://github.com/kostas1515/iif}
\end{abstract}


\section{Introduction}
\IEEEPARstart{G}{reat} advancements have been made in the field of image recognition due to deep learning techniques and the use of massive parallel computer systems. As a result, amazing technologies have been developed in the fields of automation, medicine, transportation and internet of things that make human life better. Most of these technologies use a large amount of data in order to train a deep convolutional neural network that solves the problem at hand. Even though this technique is efficient, it relies heavily on the availability of data. Models trained with curated balanced datasets like CIFAR \cite{krizhevsky2009learning}, ImageNet \cite{deng2009imagenet} and COCO \cite{lin2014microsoft} achieve good performance in many image recognition tasks such as classification, object detection and instance segmentation. However, in the real world the data are rarely balanced, instead they follow a long-tailed distribution \cite{liu2019large}, i.e. the data are imbalanced and not uniform, resulting in a major performance degradation  \cite{gupta2019lvis,li2020overcoming,wang2020devil}.   In essence, the models trained with long-tailed data can recognise the frequent (head) classes of the dataset but they fail to recognise the rare (tail) classes. As a consequence, the models that disregard the long-tailed nature of the problem, become unreliable and might raise serious concerns in critical scenarios (e.g. autonomous driving).

\begin{figure}
    \centering
    \includegraphics[scale=0.2]{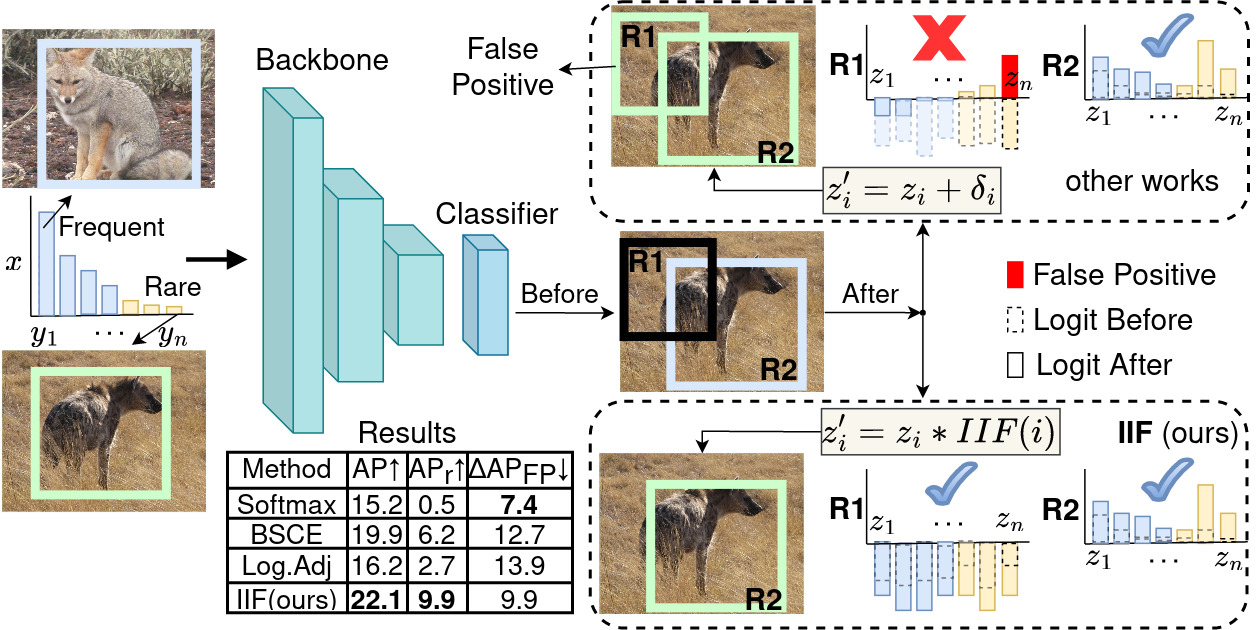}
    \caption{Previous works that use additive margin adjustment may produce a lot of false positives when used in long-tailed instance segmentation. Given a background proposal R1 which has negative logits, an additive margin adjustment may alter the sign of logits into being positive and confuse the background class for rare classes, as shown in the top branch. In the bottom branch, we propose a multiplicative $IIF$ adjustment that keeps the sign of the original predictions unchanged, thus producing less false positives while debiasing the model in favour of rare classes. In this way the model achieves better overall $AP$, rare category $AP_r$ and makes less False Positives $\Delta AP_{FP}$ than previous works such as BSCE \cite{NEURIPS2020_2ba61cc3} and Log. Adj \cite{menon2021longtail}.}
    \label{fig:iif_motivation}
\end{figure}

The cause of the performance drop in long-tailed datasets is classification imbalance \cite{oksuz2020imbalance,wang2020devil}. In detail, the frequent classes of those datasets dominate the training procedure and the network learns more about them and less about the rare classes. 
One way to solve this problem is to collect more samples from the rare categories so that in the end the data distribution will be balanced. Unfortunately, this solution costs a lot of effort and it cannot address the issue completely, as the more samples one gathers, the more categories will appear making the annotation procedure intractable. For example, if one wants to gather more images of a rare class i.e., ``remote control'' object, then one should also annotate the ``television'' object and perhaps all other objects that appear inside the living room scene that will have a higher frequency than the ``remote control''. This is a natural phenomenon of our physical world that the object frequencies follow the Zipf's law \cite{liu2019large}, making the class distribution long-tailed.

Recent approaches tackle long-tailed classification by improving the classification layer of the model \cite{NEURIPS2020_2ba61cc3,menon2021longtail,hong2021disentangling,zhang2021distribution,Kang2020Decoupling,9081988,iscen2021cbd,7780949,wang2017learning,khan2017cost,cui2019class,cao2019learning}. Margin adjustment techniques like \cite{NEURIPS2020_2ba61cc3,menon2021longtail,hong2021disentangling,zhang2021distribution,Kang2020Decoupling,cao2019learning}, are popular and intuitive classifier learning methods that have a strong theoretical foundation in label distribution shift and demonstrate performance. 

However, most of margin adjustment techniques use an additive hand-crafted margin \cite{menon2021longtail,NEURIPS2020_2ba61cc3,hong2021disentangling,cao2019learning} that is suitable for image classification but falls short of downstream tasks such as long-tailed instance segmentation because of the background class. For example, given a background object proposal, the additive margin adjustment may force the rare class logits of the background proposal to become positive.  Consequently, false positive detections will be produced by confusing the background class for a rare class as shown in the top branch of Figure \ref{fig:iif_motivation}. In contrast to this, our method uses a multiplicative margin, which only amplifies the original logits, without changing their sign, thus it does not produce false positives as shown in the bottom branch of Figure \ref{fig:iif_motivation}. 
This problem does not apply in long-tailed classification because all categories are foreground and the trade-off exists only between frequent and rare classes. In long-tailed instance segmentation however, a trade-off exists between both foreground and background classes and between frequent and rare categories.

To make this concrete, we use the TIDE toolkit \cite{tide-eccv2020} to measure the false positive detections and average precision ($AP$) \cite{everingham2010pascal} of popular margin-adjustment methods like Logit Adjustment\cite{menon2021longtail} and Balanced Softmax \cite{NEURIPS2020_2ba61cc3}. TIDE has some limitations, i.e., its error metrics do not complement $AP$ ($\Delta AP + AP \neq 1$). Nevertheless, it is useful for comparing the relative errors among models.
TIDE breaks down the error into many types such as classification, localisation and miss-detection. In this analysis, we use only $\Delta AP_{FP}$ which is the $AP^{50}$ performance loss due to false positives.  As shown in Figure \ref{fig:iif_motivation}, Softmax has low $AP$, because it fails to detect rare categories, i.e., it has low $AP_r$. Hand-crafted margin techniques like Logit Adjustment \cite{menon2021longtail} and Balanced Softmax (BSCE) \cite{NEURIPS2020_2ba61cc3} boost the performance of rare classes but they make a lot of false positives as they have increased $\Delta AP_{FP}$ compared to Softmax. 

There are a few ways to reduce false positives in long-tailed instance segmentation. Recent works \cite{zhang2021distribution,Kang2020Decoupling,NEURIPS2020_2ba61cc3,wang2021seesaw} calculate learnable margin transformations during two-stage learning. However, their margins cannot not be easily explained and it requires additional training resources to learn them. Other works, disentangle foreground from background classes by introducing an objectness branch or use zero margin for the background class. However, it is difficult to find a suitable margin for the background class, as this depends on the architecture of the detector (i.e., two-stage vs one-stage) and it cannot be calculated from the dataset.

Motivated by this, we develop a strong dataset-dependent margin adjustment technique called Inverse Image Frequency ($IIF$). Our $IIF$ uses a multiplicative adjustment, thus it reduces false positives compared to additive adjustment methods, as it only amplifies the original predictions keeping their sign unchanged, as shown in Figure \ref{fig:iif_motivation}, bottom branch. Moreover, our vanilla $IIF$ method has the best instance segmentation performance and produces less false positives as it achieves lower $\Delta AP_{FP}$, compared to similar margin adjustment techniques like Balanced Softmax (BSCE) \cite{NEURIPS2020_2ba61cc3} and Logit Adjustment (Log. Adj.) \cite{menon2021longtail}.

At the same time, it achieves strong performance on long-tailed image classification reaching $55.8\%$ top-1 accuracy on ImageNet-LT when using ResNet50 \cite{he2016deep} backbone surpassing the state-of-the-art methods by up to $3\%$. Moreover, it outperforms the state-of-the-art methods in the long-tailed instance segmentation LVIS benchmark \cite{gupta2019lvis} boosting the rare category performance by $17.5\%$ compared to vanilla Softmax.

We describe our contributions as follows:
\begin{itemize}
    \item We show that previous handcrafted margin adjustment techniques used in classification may produce false positives in long-tailed instance segmentation as a result of background class.
    \item We develop a robust margin adjustment method $IIF$ that boosts the performance of rare categories and makes fewer false positive detections compared to other margin adjustment methods.
    \item We evaluate our $IIF$ method on CIFAR10-LT, CIFAR100-LT, ImageNet-LT, Places-LT, LVISv1 and we show that it surpasses the state-of-the-art methods by a significant margin.
\end{itemize}

\section{Related Work}
\label{subsec:img_cls}
Long-tailed image recognition has received a lot of interest in recent years and many works have been developed, a summary of them can be found in these surveys \cite{yang2022survey,oksuz2020imbalance}. Many long-tailed datasets have been created for object classification \cite{cao2019learning,liu2019large}, scene classification \cite{liu2019large,wang2017learning}, species classification \cite{van2018inaturalist}, faces recognition \cite{liu2022open,robinson2023balancing}, object detection \cite{gupta2019lvis,kuznetsova2020open} and instance segmentation \cite{gupta2019lvis}. Recently, more datasets \cite{yang2022multi,tang2022invariant,gu2022tackling} and works \cite{jamal2020rethinking,jing2021towards} were proposed that tackle both the long-tailed and domain adaptation problem simultaneously. The datasets are created either by extracting datasets from the wild, or by sub-sampling balanced datasets. These datasets can be characterised by their imbalance factor $\beta={n_{max}}/{n_{min}}$ which is the ratio between the maximum and minimum class frequency on the training set. As shown in Table \ref{tab:imb_factor}, the most imbalanced dataset is LVIS \cite{gupta2019lvis} and the least imbalanced is CIFAR-LT \cite{cao2019learning}. Note that COCO \cite{lin2014microsoft} is artificially balanced in the sense that all classes have a large and diverse set of images. However, COCO has a larger imbalance factor than common long-tailed classification datasets as the class frequencies are not uniform. This is due to the fact that COCO is a densely annotated scene-centric dataset and this makes it difficult to have totally balanced classes due to the Zipfean distribution. Regarding the testing set in these datasets, most of them adopt a balanced test set, so that the performance is evaluated fairly on all categories. For the case of object detection, it is difficult to have balanced test set, due to scene-centric images. Despite that, when using mAP, every category is evaluated independently and has equal contribution to the final performance.
\begin{table}
    \centering
    \caption{Characteristics of image recognition datasets. Table adjusted from \cite{yang2022survey}}
    \begin{tabular}{c|c|c|c}
    \hline
         Dataset&$\beta$&Train Distribution & \# of Images \\
         \hline
         CIFAR-LT \cite{cao2019learning}&100&Exponential&50K\\
         ImageNet-LT\cite{liu2019large}&256&Pareto&186K\\
         Places-LT\cite{liu2019large}&996&Pareto&62.5K\\
         COCO\cite{lin2014microsoft}&1,325&Balanced&118K\\
         LVISv1\cite{gupta2019lvis}&50,552&Long-tailed&99K\\
    \hline
    \end{tabular}
    \label{tab:imb_factor}
\end{table}

\begin{table}
    \centering
    \caption{Related Works}
    \begin{tabular}{c|c|c}
    \hline
         Family& Method&Reference \\
         \hline
         \multirow{6}{*}{\shortstack[l]{Representation \\ Learning}}& Re-sampling&\cite{chawla2002smote,mahajan2018exploring,park2022majority,shen2016relay,Kang2020Decoupling,gupta2019lvis}\\
         &Distillation&\cite{iscen2021cbd,He_2021_ICCV}\\
         &Feature Generation&\cite{wang2021rsg,vigneswaran2021feature,zang2021fasa,hong2022safa}\\
         &Contrastive Learning&\cite{wang2021contrastive,samuel2021distributional}\\
         &Fusion &\cite{wang2020long,zhou2020bbn,guo2021long}\\
         &Data Augmentation &\cite{zhang2017mixup,yun2019cutmix,szegedy2016rethinking,zhang2021delving,cubuk2019autoaugment,cubuk2020randaugment}\\
         \hline
         \multirow{4}{*}{\shortstack[l]{Classifier \\ Learning}}& Cost-sensitive Loss&\cite{7780949,wang2017learning,khan2017cost,cui2019class}\\
         &Gradient Balancing&\cite{tan2020equalization,tan2021eqlv2,hsieh2021droploss,wang2021seesaw,wang2021adaptive,Feng_2021_ICCV,alexandridis2022long}\\
         &Two-Stage Methods&\cite{Kang2020Decoupling,9081988,wang2020devil,hsu2023abc,zhang2021distribution}\\
         &Margin Adjustment&\cite{NEURIPS2020_2ba61cc3,menon2021longtail,cao2019learning,zhang2021distribution,hong2021disentangling,Kang2020Decoupling,ye2020identifying}\\
         \hline
    \end{tabular}
    \label{tab:related_works}
\end{table}

Many solutions have been developed inside the long-tailed paradigm and they can be categorised in representation learning and classifier learning techniques as shown in Table~\ref{tab:related_works}. 

\subsection{Representation Learning}
A simple representation learning technique is to re-sample the data distribution \cite{chawla2002smote,mahajan2018exploring,park2022majority,shen2016relay,Kang2020Decoupling,gupta2019lvis}, by either oversampling or downsampling the classes of datasets. Despite having satisfactory performance, oversampling requires additional computing resources and may cause overfitting for tail classes while undersampling does not exploit efficiently the available data and may cause underfitting for head classes.

Other representation learning techniques enhance the quality of the deep feature extractor by using contrastive learning and supervised learning \cite{wang2021contrastive,samuel2021distributional}. However, such methods require a laborious multi-stage training pipeline or the construction of multi-branch networks in order to combine the supervised and contrastive objectives effectively. 

Some techniques enhance the feature extractor by generating rare category samples \cite{wang2021rsg,vigneswaran2021feature,zang2021fasa,hong2022safa}. However, generated features are usually perturbed versions of the old features thus they improve the quantity, rather than the quality of features.
In addition to this, distillation methods \cite{iscen2021cbd,He_2021_ICCV} have been proposed to efficiently exploit the representation quality of larger capacity models. These methods have good results, but they require additional training resources for learning the teacher models.
Fusing methods use a two-branch network trained with random and oversampling strategy \cite{zhou2020bbn,guo2021long} or learn ensemble models \cite{wang2020long} that specialise in rare and frequent categories. They have shown good performance but it is at expense of additional training resources. 
Finally, data augmentation methods such as mixup \cite{zhang2017mixup}, cutmix \cite{yun2019cutmix}, label smoothing \cite{szegedy2016rethinking,zhang2021delving} and AutoAugment \cite{cubuk2019autoaugment,cubuk2020randaugment} improve the generalisation ability of the model for all classes.

\subsection{Classifier Learning}
\subsubsection{Cost-sensitive Learning} ~\cite{7780949,wang2017learning,khan2017cost,cui2019class,liu2017cost} methods assign costs to samples according to the dataset's distribution in order to balance the training and learn all classes. They can produce good results without the need of extra training resources but they require careful calibration, hyper-parameter tuning and they are difficult to design and optimise as the costs may be excessive and destabilise training.

\subsubsection{Gradient Balancing} \cite{tan2020equalization,tan2021eqlv2,hsieh2021droploss,wang2021seesaw,wang2021adaptive,Feng_2021_ICCV} methods assign weights to the gradients produced by positive and negative samples, or use different activation functions for gradient balancing \cite{alexandridis2022long}. These techniques are most useful in long-tailed object detection and long-tailed instance segmentation as in such tasks the special background class magnifies the imbalance and increases the complexity of the task. 

\subsubsection{Two-stage Techniques}  \cite{Kang2020Decoupling,9081988,wang2020devil,hsu2023abc,zhang2021distribution} first optimise the model to classify the head classes and in the latter stage, they finetune or retrain it for the rare classes. This is achieved using re-sampling techniques, weight normalisation techniques or transfer learning so that in the end the model can classify both head and tail classes effectively. This technique can alleviate the bias of the classifiers and it is task agnostic. Nevertheless, it may require the construction of a complex pipeline and additional training resources. 

\subsubsection{Margin Adjustment} \cite{NEURIPS2020_2ba61cc3,menon2021longtail,cao2019learning,zhang2021distribution,hong2021disentangling,Kang2020Decoupling,ye2020identifying} alter the decision boundary of the classifier either during training or a posterior to shift the predicted distribution. The resulting classification boundary is closer to the head classes and further away from the tail classes and the feature space of head classes becomes smaller while the space of tail classes is enlarged. This way, during inference the adjusted classifier is less biased towards predicting the head classes. 

The margin adjustment techniques produce good results, but they have limitations. For example, \cite{NEURIPS2020_2ba61cc3,menon2021longtail,hong2021disentangling,cao2019learning} use an additive adjustment for long-tailed image classification but this may produce many false positives in downstream tasks such as long-tailed instance segmentation, as they do not explicitly model the background class margin. Moreover, learnable margin transformation techniques \cite{Kang2020Decoupling,zhang2021distribution} require a two-stage strategy and therefore additional computing resources. They alleviate false positives in downstream tasks as they learn foreground and background category margins simultaneously but their margins are difficult to explain.

In contrast to these, our $IIF$ uses dataset-dependent margins that are easy to explain and use in both long-tailed image classification and long-tailed instance segmentation.

\section{Preliminaries}
\label{sec:preliminaries}
$IIF$ is closely related to ideas from label distribution shift, we follow a similar analysis as in \cite{menon2021longtail,zhang2021distribution,NEURIPS2020_2ba61cc3}. Let $p_s(y|x)$  and $p_t(y|x)$ be the source and target distributions respectively. By using the Bayes theorem, the source distribution can be written as:
\begin{equation}
    p_s(y|x) = \frac{p_s(x|y)p_s(y)}{p_s(x)}
    \label{eq:p_source}
\end{equation}
and the target distribution can be written as:
\begin{equation}
    p_t(y|x) = \frac{p_t(x|y)p_t(y)}{p_t(x)}
    \label{eq:p_target}
\end{equation}
If one assumes that the data generating functions are equal $p_s(x|y) = p_t(x|y)$ then by dividing Eq. \ref{eq:p_source} and Eq. \ref{eq:p_target}, one can rewrite the target distribution as:
\begin{equation}
    \begin{split}
        \frac{ p_s(y|x) }{p_t(y|x)} = & c(x) \frac{p_s(y)}{p_t(y)} \\
        p_t(y|x) = & \frac{1}{c(x)}\frac{p_t(y)}{p_s(y)} p_s(y|x)
    \label{eq:dist_shift}
    \end{split}
 \end{equation}
where $c(x)=\frac{p_t(x)}{p_s(x)}$.
During training $p_s(y|x)$ is approximated by the model $f_{y}(x;\theta)$ and a scorer function $s(x)=e^x$:
\begin{equation}
    p_s(y|x) \propto e^{f_{y}(x;\theta)}
    \label{eq:soft_approx}
\end{equation}
By using Eq. \ref{eq:dist_shift} and Eq. \ref{eq:soft_approx} one can compensate for label distribution shift using the following Equation:

\begin{equation}
\begin{split}
    p_t(y|x) \propto &  \frac{1}{c(x)}\frac{p_t(y)}{p_s(y)} e^{f_{y}(x;\theta)}\\
    =  & e^{f_{y}(x;\theta) + \log(p_t(y)) - \log(p_s(y)) - \log(c(x))}
    \label{eq:pc_softmax}
\end{split}
\end{equation}
During inference, one is interested to predict a single class $\bar{y}$ and this is usually achieved by taking the maximum value of Eq. \ref{eq:pc_softmax}:
\begin{equation}
    \begin{split}
    \bar{y}= \argmax_y(f_{y}(x;\theta) + \log(p_t(y))\\ 
    -\log(p_s(y)) - \log(c(x))) 
    \end{split}
    \label{eq:inference_softmax}
\end{equation}
Moreover, one can simplify Eq. \ref{eq:inference_softmax} by eliminating $c(x)$ because it is invariant to $\argmax_y$ as follows:
\begin{equation}
    \begin{split}
    \bar{y}= \argmax_y(f_{y}(x;\theta) + \log(p_t(y))\\ 
    -\log(p_s(y)) ) 
    \end{split}
    \label{eq:label_shift_general}
\end{equation}
Using Eq. \ref{eq:label_shift_general} one can solve the label distribution shift problem. However, in the real world, $p_s(y)$ and $p_t(y)$ may be unknown. Luckily, one can still solve the label shift problem by estimating $p_s(y)$ and $p_t(y)$ from the data. 
\subsection{Training-Set Distribution}
\label{subsec:train_set_assumption}
First, even though $p_s(y)$ is unknown, one has access to a training set $D$ that is sampled uniformly from the source distribution $s$. Thus, instead of calculating $p_s(y)$ one can use $p_D(y)$, which is the class distribution on the training set. Accordingly, as $|D|$ grows larger then one can be more certain that $p_D(y)$ will be a good estimate of $p_s(y)$.
\subsection{Test-Set Distribution}
\label{subsec:test_set_assumption}
Generally, $p_t(y)$ can be any arbitrary distribution and when $p_t(y) \neq p_s(y)$ there exists label shift. If the label shift is unknown, i.e. $p_t(y)$ is not known, then it can be estimated using the model's predictions as suggested in \cite{lipton2018detecting}. In the case of long-tailed image recognition, the target distribution is uniform because the test set is balanced. The reason is that, in long tailed visual benchmarks, every category is evaluated fairly and it contributes equally to the final performance \cite{liu2019large,cao2019learning,menon2021longtail,NEURIPS2020_2ba61cc3}.  Therefore, $p_t(y)= \frac{1}{C}$ where $C$ is the total number of classes in the dataset.

To this end, we can rewrite Eq. \ref{eq:label_shift_general} as:
\begin{equation}
     \bar{y}= \argmax_{y}(f_{y}(x;\theta) - \log(p_D(y)))
    \label{eq:pc_softmax_simple}
\end{equation}
In essence, Eq. \ref{eq:pc_softmax_simple} suggests that one can compensate for this label distribution shift by translating the model's output $f_{y}(x;\theta)=z$ by the training set's class probability:
\begin{equation}
    z'=z-\log(p_D(y))
    \label{eq:additive_iif}
\end{equation}
\subsection{Limitations}

Despite that, in downstream tasks like instance segmentation or object detection, there is also the special background class $b$, that depends on the model's configuration, i.e., one-stage detectors \cite{redmon2018yolov3,lin2017focal} have a different estimate of $b$ than two-stage detectors \cite{ren2015faster,he2017mask} that use region proposals. The background class is usually handled by predicting C+1 categories using softmax, but in this way, Eq. \ref{eq:additive_iif} is not directly applicable as $p_D(y=b)$ cannot be easily calculated. Additionally, if a bad background probability estimate is used, this may cause false positives and deteriorate the model's performance.

Some works \cite{Kang2020Decoupling,zhang2021distribution} alleviate this problem by learning the foreground and background class margins during two-stage learning but these margins are difficult to explain, and may cause concerns in safety critical applications. Other works like \cite{wang2021seesaw} use an objectness branch to reduce false positives. They predict two extra logits that determine whether the sample belongs to foreground and background respectively. This technique disentangles the classification task to two sub-tasks, i.e. background and foreground prediction; in this way, foreground class margins can be applied easily to foreground samples.  However, the usage of objectness branch hurts the model's Fixed-AP performance \cite{dave2021evaluating}, as it only improves the cross category rankings as suggested by \cite{dave2021evaluating}. Recently, Hsieh et al. \cite{hsieh2021droploss} studied the background category problem and propose DropLoss, a loss that assigns weights to background gradients in an adaptive manner. However, they utilised a gradient re-balancing method which is different from margin adjustment techniques.

For these reasons, we develop $IIF$ using dataset-dependent margins that are easy to explain and use in long-tailed classification and long-tailed instance segmentation. Our $IIF$ alleviates for label distribution shift in long-tailed benchmarks. At the same time it uses a multiplicative adjustment, that keeps the original sign of the predictions unchanged thus it reduces the false positive detections compared to other additive margin adjustment methods. 

\section{Methodology}

\subsection{Inverse Image Frequency}
Inverse Image Frequency (IIF) is inspired by Inverse Document Frequency (IDF). IDF is an important heuristic method that reweights textual terms according to their relevance and it has been extensively used in text retrieval tasks \cite{seki2002sentence,sajton1988term,8370597,paik2013novel}. IDF reweighs a term according to the number of documents the term appears in the corpus. In our work, instead of measuring the number of documents where a term appears, we measure the number of images where an object appears.

In detail, given a set of training images $D$ sampled from the source distribution $s$, Image Frequency $IF(y,D)$ of a class $y \in \mathbb{N}$ is computed as the number of images in which an object $o_{y}$ appears:
\begin{equation}
    IF(y,D) = |\{image \in D: o_{y} \in image\}|
\end{equation}
The class probability $p_D(y)$ of class $y$ is defined as:
\begin{equation}
    p_D(y)=\frac{IF(y,D)}{K}
\end{equation}
where $K=\sum_{y=1}^{y=C}IF(y,D)$ and $C$ is the total number of classes in $D$. Next, $IIF$ is measured by taking the logarithm of the inverse of $p(y)$\footnote{We omit $D$ for simplicity since all following calculations are performed in the training set $D$.}, i.e.,
\begin{equation}
\label{eq:raw_idf}
    IIF(y)=  \log\frac{K}{IF(y)} = -\log(p(y))
\end{equation}
Next, one can transform the logits $z$ of the classification layer using the $IIF$ transformation.
\begin{equation}
    z_{IIF}=-z  \log(p(y))
    \label{eq:multiplicative_iif}
\end{equation} 
This feature transformation is similar to the IDF feature transformation whose justification is explained in \cite{robertson2004understanding}.

The use of the logarithm is convenient because it links the probability space $(0,1)$ to real space enhancing the compatibility of the predicted logits $z$ and $IIF$ weights. Other link functions are discussed in Table~\ref{tab:link_functions}.

When $IIF$ is multiplied with the logits of the classification layer, it redistributes the weights across different classes.  The weights of $IIF$ are larger for the rare classes than the frequent classes thus, it can be used to remove the frequent category bias and alleviate class imbalance.

The  Eq.~\ref{eq:multiplicative_iif} resembles Eq. \ref{eq:additive_iif}. Its difference is that instead of additive adjustment, it performs multiplicative adjustment. The multiplicative adjustment benefits both long-tailed classification and long-tailed instance segmentation as it alleviates class imbalance and it makes fewer false positive detections since it maintains the sign of the original predictions intact. 
If the detector predicts logits that are negative for one background region, then an additive adjustment may force them inside the detection threshold, making the model overconfident and producing false detections. In contrast to that, the multiplicative adjustment will only amplify the logits, in other words, it will not affect their sign and keep background predictions outside the detection threshold as shown in the bottom branch of Figure \ref{fig:iif_motivation}.

\subsubsection{Connection to Softmax}  In practice, neural networks typically produce a probabilistic vector $q$ by using a softmax output layer $\sigma$. This converts the logit $z_i$ for each class $i$ into a probability $q_i$, by comparing $z_i$ with the other logits $ q_i = \frac{\exp({z_i})}{\sum_{j=1}^{C}{\exp({z_j})}}$. 

The dominant prediction $q_i$ can be found by computing $\argmax_{i\in C}z_i$, and this holds true because all $z_i$ are activated by the same strictly increasing activation function $f(x)= e^x$. Changing the base of the activation function from $e$ to any $\alpha \in {\rm I\!R}$ would not affect the ranking and this is fair for balanced datasets. For imbalanced datasets, we can change the base of the activation function for each $z_i$ according to the inverse class probability i.e., $f^i(x)= (\frac{1}{p(i)})^x$ and compensate for imbalance. In this way, we allow logits that correspond to rare classes to get easily activated. 
We can achieve this by applying $IIF$ transformation Eq. \ref{eq:multiplicative_iif}:
\begin{equation}
    \begin{aligned}
    q_{IIF,i}
    &=\frac{\exp({z_i\log{(\frac{1}{p(i)})}})}{\sum_{j=1}^C{\exp({z_j\log{(\frac{1}{p(j)})}})}}\\
    &=\frac{(\frac{1}{p(i)})^{z_i}}{\sum_{j=1}^C(\frac{1}{p(j)})^{z_j}}
    \label{eq:iif_softmax}
    \end{aligned}
\end{equation}
Note that, here the class index starts from 1 to C, but in the case of instance segmentation there exist a background class $b$ that is usually encoded as the ``0'' class. In this case, there could be C+1 classes in softmax and the index starts from 0 to C.

Equation \ref{eq:iif_softmax} has two beneficial properties. First, it maintains the property that $\sum_{i}^Cq_{IIF,i}=1$, this means that $q_{IIF,i}$ is a probabilistic vector, the proof is provided in Appendix. Secondly, re-balancing occurs naturally, as logits $z_i$ corresponding to frequent classes i.e., $p(i) \xrightarrow[]{}1 $ will not contribute as much in softmax because they will be irrelevant, $(\frac{1}{p(i)})^{z_i} \xrightarrow[]{}1,\forall z_i$. On the other hand, logits $z_i$ corresponding to rare classes i.e., $p(i) \xrightarrow[]{}0 $, will determine the final outcome of softmax $(\frac{1}{p(i)})^{z_i} \xrightarrow[]{} +\infty$.

To make the second point concrete, one can consider an extreme example of binary classification where $p(y=0)=0.99999$ and $p(y=1)=0.00001$. $IIF$ will significantly downgrade $z_0$ rendering it irrelevant and it will make $z_1$ the dominant factor in softmax. In other words, $IIF$ will make softmax more sensitive to $z_1$ than $z_0$ which is the class that matters most in this hypothetical example. Compared to previous works that perform additive adjustment, $IIF$ makes stronger adjustment, because of the multiplicative function, and it enlarges the rare class probabilities with a faster rate than the additive case.

\subsubsection{Variants}
Moreover, one can define $IIF$ variants by using different log bases or different link functions in order to transform the probability space into the real space. The motivation is that the imbalance factor changes according to the dataset thus, it may be beneficial to use different variants that provide stronger debiasing effects.

In Table~\ref{tab:link_functions}, some basic variants are summarised and in Figure~\ref{fig:link_functions} their behaviour is illustrated. 
\begin{itemize}
    \item The raw $IIF$ is the most straightforward way to transform the probabilities into weights. The different log-bases can be used to control the magnitude of the weights when dealing with very low probabilities. 
    \item The smooth $IIF$ has similar behaviour to raw $IIF$, but it has the advantage of handling zero image frequency values, thus it can be used either on the full training dataset $D$ or on the mini-batch $d$ in online fashion using the mini-batch statistics.
    \item The relative $IIF$ uses the inverse logit link function and it has a bigger range of values than the smooth or raw $IIF$. It is a symmetrical function around 0.5 and it is useful when modelling binary events. In the long tail scenario, usually most class probabilities are below 0.5 thus the relative link will produce only positive weights and will have similar behaviour with the raw $IIF$.
    \item The Normit $IIF$ assumes that the data follow Gaussian Distribution. It is has similar properties to the relative $IIF$, it is also symmetrical around 0.5, but it has a smoother slope.
    \item The Gombit $IIF$ assumes that the data follow Gompertz Distribution. It uses an asymmetrical link function that puts more emphasis to small probability events as it produces increasingly larger positive weights compared to high probability events. In other words, the growth rate for the response value is larger as the probability gets smaller.
\end{itemize}

In addition to Inverse Image Frequency, one can calculate Inverse Object Frequency $IOF$ by counting object instances instead of images. In tasks such as image classification, $IIF$ and $IOF$ will produce the same result as objects and images have a one-to-one relationship. For other tasks such as instance segmentation, multiple objects can coexist in a single image thus the two methods are different and they produce different weights.

\begin{table*}[t]
\begin{minipage}{0.5\linewidth}
\centering
\caption{Inverse Image frequency variations, $\Phi$ denotes the inverse cumulative distribution function of Normal distribution}
    \begin{tabular}{c|c|c}
    \hline
         $IIF$ &Formula&Range  \\
         \hline
         Raw&$\log\frac{K}{IF_y}$&$(0,\infty)$\\[3pt]
         
         Smooth&$ \log\frac{K+1}{IF_y +1} +1$&$[1,\infty)$ \\[3pt]
         
         Relative& $\log\frac{K-IF_y}{IF_y}$ &$(-\infty,\infty)$\\[3pt]
         
         Base $10$& $\log_{10}\frac{K}{IF_y}$&$(0,\infty)$ \\[3pt]
         
         Base $2$& $\log_{2}\frac{K}{IF_y}$&$(0,\infty)$ \\[3pt]
         
         Gombit& $-\log-\log(1-\frac{IF_y}{K})$&$(-\infty,\infty)$ \\[3pt]
         
         Normit& $\Phi^{-1}(\frac{K}{IF_y})$&$(-\infty,\infty)$ \\[3pt]
         \hline
    \end{tabular}
    
    \label{tab:link_functions}
\end{minipage}
\begin{minipage}{0.5\linewidth}
\centering
\includegraphics[scale=0.37]{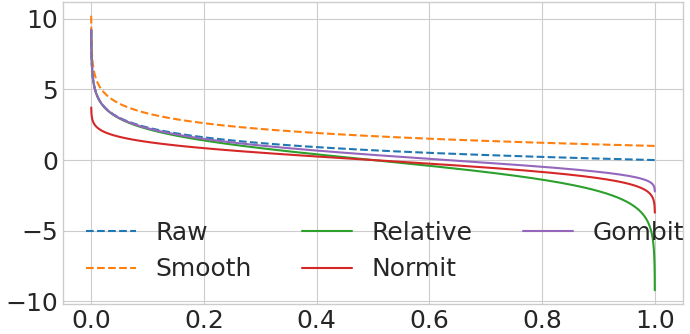}
\captionof{figure}{Inverse Image Frequency curves. Variations of Table \ref{tab:link_functions} are illustrated. The x-axis denotes the input and the y-axis denotes the output of $IIF$. }
\label{fig:link_functions}
\end{minipage}
\end{table*}

\subsubsection{IIF Cross Entropy}
$IIF$ can be integrated during training by optimising the $IIF$ Cross Entropy loss. Let $\mathcal{Y}$ be the ground truth one-hot encoded vector of class $y$ then by using Eq. \ref{eq:iif_softmax} the loss is:
\begin{equation}
\begin{aligned}
    CE_{IIF}(q,\mathcal{Y}) &= -\sum_{i=1}^{C}\mathcal{Y}_i\log ( q_{IIF,i} ) \\
    &=-\log\frac{\exp(z_y\log{(\frac{1}{p(y)})})}{\sum_{j=1}^{C}\exp(z_j\log{(\frac{1}{p(j)})} )} \\
    &=-z_y \log(\frac{1}{p(y)})  + \log({\sum_{j=1}^{C} (\frac{1}{p(j)})^{z_j} })
    \label{eq:ce_iif}
\end{aligned}
\end{equation}

It can be seen that when the class probability is higher i.e., $p(y)\xrightarrow[]{}1$, then there is loss only for the negative classes and the network does not receive any information about the target class $y$. On the other hand, when the class probability is low, i.e $p(y)\xrightarrow[]{}0$, then the positive class $y$ dominates the loss, forcing the model to focus on the rare class. In the end, this allows the model to learn more about the categories whose class probability is low. 

For long-tailed image classification, all classes are foreground and $IIF$ can be used without modifications. That is not the case for long-tailed instance segmentation as there exist background samples. 

In instance segmentation, many models encode the background samples as the ``0'' class. Thus they predict a logit vector $z=[z_0,z_1,...,z_C]$. To apply our $IIF$ in this case, we set the background weight as 1, i.e. $IIF=[1,-\log(p(1)),...,-\log(p(C))]$, to keep the background object's estimation unaltered and only change the foreground objects' estimations.

Using $IIF$ Cross Entropy Loss Eq. \ref{eq:ce_iif}, the gradient is shown in Eq. \ref{eq:iif_ce_grad}. The proof is provided in appendix.
\begin{equation}
    \frac{\partial{CE_{IIF}}}{\partial{z_i}}=
    \left\{
	\begin{array}{ll}
		-\log(p(i))(q_{IIF,i}-1)  & \mbox{if } i = y \\
		-\log(p(i))q_{IIF,i} & \mbox{otherwise } 
	\end{array}
\right.
\label{eq:iif_ce_grad}
\end{equation}
It can be seen that the positive gradient, i.e., when $i=y$ will be larger in magnitude when the class probability $p(i)$ of the target is low. This will encourage the model to learn more about the rare classes of the dataset. Additionally, the negative gradients i.e., when $i \neq y$, will be weighted according to their class probabilities. This means that negative gradients occurring from frequent categories will be suppressed. In the end, using $IIF$ the model becomes more sensitive to rare classes as their gradients will be upweighted.

\subsubsection{Post-process IIF}
\begin{figure}
    \centering
    \includegraphics[scale=0.35]{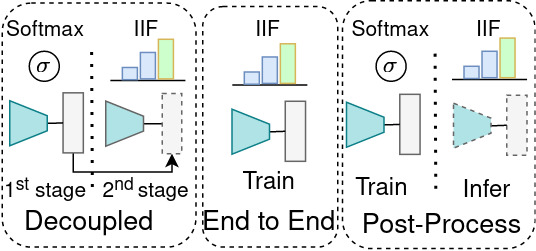}
    \caption{$IIF$ strategies. Left:  $IIF$ can be used in decoupled strategy, where first the whole model is trained with Softmax and in the second stage only the classifier is retrained using $IIF$ Cross Entropy (Eq. \ref{eq:ce_iif}). Middle: the whole model is trained with $IIF$ Cross Entropy (Eq. \ref{eq:ce_iif}). Right: the model is trained with Softmax and only during inference $IIF$ weights are injected in the model's predictions using Eq.~\ref{eq:iif_softmax} as post-processing method.}
    \label{fig:iif_strategies}
\end{figure}

Moreover, $IIF$ can be applied during inference using Eq. \ref{eq:iif_softmax} as a post-processing method. If post-processing $IIF$ is applied then it is no longer necessary to use Eq. \ref{eq:ce_iif}. Instead, vanilla Cross Entropy can be used to train a model and only during inference Eq. \ref{eq:iif_softmax} can be injected into the model's output in order to de-bias the predictions. In conclusion, all $IIF$ strategies can be illustrated in 
Fig. \ref{fig:iif_strategies}.

\subsection{IIF Cross Entropy versus Cost Sensitive Learning} 
$IIF$ Cross Entropy re-weights the gradient of each sample $i$ according to its Inverse Image Frequency $-\log(p(i))$, as shown in Eq.~\ref{eq:multiplicative_iif}.
This differs from Cost-Sensitive Learning method (\textit{CSL}) that re-weights all samples based on scalar weight $\alpha_y$. In principle, \textit{CSL} applies weight multiplication to the loss rather than the logits, more details on \textit{CSL} can be found in this work \cite{cui2019class}. The gradient of \textit{CSL} for a sample $i$ is:
\begin{equation}
    \frac{\partial{L_{CSL}}}{\partial{z_i}}=
    \left\{
	\begin{array}{ll}
		\alpha_y(q_i-1)  & \mbox{if } i = y \\
		\alpha_yq_i & \mbox{otherwise} 
	\end{array}
\right.
\end{equation}

To better understand how \textit{CSL} differs from $IIF$, we can set $\alpha_y=-\log(p(y))$ and assume that $q_i=q_{IIF,i}$. The positive gradient of \textit{CSL}, (i.e. $i=y$), will be the same with $IIF$ whereas the negative will differ. In \textit{CSL}, the negative gradient will be multiplied by the scalar $-\log(p(y))$ which is the weight of the target class, whereas in $IIF$ it is multiplied by its class respective weight $-\log(p(i))$. The latter balances negative gradients more efficiently and suggests that positive and negative gradients should not be valued the same.

In practice, $CSL$ might be unstable during the early phases of training because of the imbalance between positive and negative gradients which is magnified by the weight $\alpha_y$. Consequently, it requires careful hyperparameter tuning to balance the dynamics in mini-batch training.  $IIF$ on the other hand, suppresses the imbalance caused by the negative gradients as it re-weights them based on their class probabilities.

\subsection{Connection to Other Works}
\label{sub:hiif_variant}
$IIF$ has a similar idea to recent successful margin adjustment techniques shown in Table \ref{tab:sim_adj_methods} as all of these methods reweight the logits based on either probabilities, image frequencies or learnable weights. 
The important detail of $IIF$ is that it does multiplicative adjustment, thus it makes fewer false positives than additive handcrafted margin adjustment techniques for downstream tasks. Moreover, since it is a dataset-dependent method, it is easier to interpret and justify than other learnable margin adjustment approaches.

In addition to this, our method can also be compared to calibration techniques such as Platt Scaling \cite{Platt99probabilisticoutputs}, Temperature Scaling \cite{hinton2015distilling,ye2020identifying} or NorCal \cite{pan2021model}. In comparison to \cite{Platt99probabilisticoutputs}, $IIF$ reweights the classification logits based on dataset statistics rather than learnable parameters, in contrast to \cite{hinton2015distilling} it uses class specific weights instead of a single global temperature and different from \cite{pan2021model,ye2020identifying} $IIF$ does not require additional hyperparameters.

\begin{table}
    \centering
    \caption{Margin adjustment techniques}
    \begin{tabular}{c|c|c}
    \hline
         Method&Type&Adjustment \\
         \hline
         LDAM \cite{cao2019learning}&Dataset-Dependent&$z_{i}^\prime=z_i-c/IF(i)^\frac{1}{4}$\\
         LWS \cite{Kang2020Decoupling}&Learnable& $z^\prime=\alpha_{i}z_i$\\
         Balanced Softmax \cite{NEURIPS2020_2ba61cc3}&Dataset-Dependent& $z_{i}^\prime=z_i+\log(IF(i))$\\
         Log. Adj. PostHoc\cite{menon2021longtail}&Dataset-Dependent&$z_{i}^\prime=z_i+\log(IIF(i))$\\
         Log. Adj. Loss\cite{menon2021longtail}&Dataset-Dependent&$z_{i}^\prime=z_i-\log(IIF(i))$\\
         DisAlign \cite{zhang2021distribution}&Learnable&$z_{i}^\prime=\alpha_iz_i+\beta_i$\\
         $IIF$&Dataset-Dependent&$z_{i}^\prime=IIF(i)z_i$\\
        \hline
    \end{tabular}
    \label{tab:sim_adj_methods}
\end{table}

\section{Long-tailed Classification Experiments}
\label{sec:experiment}
\subsection{Datasets and Evaluation}
In long-tailed image classification, we use CIFAR10-LT and CIFAR100-LT with exponential imbalance ratio 100 as in \cite{cao2019learning}, ImageNet-LT \cite{liu2019large} and Places-LT \cite{liu2019large} following the common long-tailed classification protocol. These datasets show a significant label shift as they have long-tailed train distribution and balanced test distribution. The balanced test distribution is artificially constructed so that the model's performance can be fairly evaluated on each class.  These datasets have the characteristics described in subsection~\ref{subsec:train_set_assumption} and~\ref{subsec:test_set_assumption} and our method can alleviate their shift from long-tailed distribution to balanced distribution.

To measure the performance of $IIF$ we use top-1 accuracy following the common evaluation protocol.

\subsection{Implementation Details}

We have observed that the standard implementation is suboptimal and can be significantly enhanced. Therefore, we create Squeeze-and-Excitation (SE) \cite{hu2018squeeze} ResNets to increase the capacity of our representation models. We choose this attention mechanism as it adds minimal complexity and has good performance. We use an SE-ResNet32 with reduction factor $r=4$ for CIFAR-LT, SE-ResNet50 and SE-ResNeXt50-4x32 with $r=16$ for ImageNet-LT. For Places-LT, we pre-train a SE-ResNet152 with $r=16$ on full ImageNet and then we finetune it according to \cite{liu2019large}. For all SE modules we use the \textit{Average} squeeze operator and the \textit{Sigmoid} excitation operator and all linear layers have the same dimensions as in \cite{hu2018squeeze} implementation. The ResNet implementation follows official Pytorch implementation \cite{paszke2019pytorch}. All models are trained using Pytorch framework and 4 Nvidia V100 GPUs.

\subsubsection{Longer Training}
We have observed that training for more epochs improves the performance of the representation model. For CIFAR-LT datasets we use a batch size of 64 and a training schedule of 400 epochs, a learning rate of 0.1 with learning rate decay at epoch 360 and 380. For ImageNet-LT, the model is trained for 200 epochs using a batch size of 256, a learning rate of 0.2 and cosine learning schedule. For Places-LT \cite{liu2019large} dataset we use an ImageNet pre-trained ResNet152 backbone. Then, we finetune its last residual block and classifier for 30 epochs using batch size 256, learning rate 0.1 and cosine scheduler.

\subsubsection{Regularisation and Augmentations}
We use cosine classifier with scale $s=16$ for ImageNet-LT and CIFAR-LT and learnable scale for Places-LT. Moreover, we use Mixup \cite{zhang2017mixup} with the factor of $0.2$ for all datasets. Regarding augmentations, we use the optimal AutoAugment \cite{cubuk2019autoaugment} policies for CIFAR-LT and ImageNet-LT and RandAugment \cite{cubuk2020randaugment} for Places-LT. 
In addition to this, we observed that the recommended weight decay used in CIFAR-LT \cite{cao2019learning} and ImageNet-LT \cite{Kang2020Decoupling} is suboptimal for our model. After conducting a grid search, we found that the value 1e-4 works well for all datasets and improves the performance.

Our findings confirm that weight decay tuning is important and should not be overlooked as also mentioned in \cite{LTRweightbalancing}.

\subsubsection{Two-stage Strategy}
Inspired by \cite{Kang2020Decoupling}, we perform experiments using the two-stage strategy when training the models with $IIF$. We use random sampling in all stages. In the first stage we pre-train the models using Softmax Cross-Entropy and in the second stage, we retrain only the classifier's weights using $IIF$. For ImageNet-LT, we use a learning rate of 2e-5 and we train the classifier for 5 epochs; for Places-LT, we use 1e-5 and we train for 10 epochs; and for CIFAR-LT, we use a learning rate of 1e-4 and we train the classifier for 20 epochs.

\subsection{Classifier Learning using $IIF$}
\subsubsection{Training Strategies}
We start our analysis by studying strategies to improve classification using $IIF$. We explore $IIF$ as decoupled strategy, as a post-processing method and as a cost-sensitive learning method.

Table \ref{tab:img_int_strat} suggests that using $IIF$ with decoupled training achieves the best performance, reaching $84.1\%$ in CIFAR10-LT, $48.9\%$ in CIFAR100-LT and $56.0\%$ in ImageNet-LT. The decoupled training is better than end-to-end training because the representations are learned more efficiently with Softmax Cross Entropy rather than other techniques as described in \cite{Kang2020Decoupling}. After learning the representations, $IIF$ can be used to retrain only the classifier and remove the frequent category bias from the model.
Moreover, $IIF$ has good results when used as a post-processing method. Under this setup, the model is first trained with Softmax and only during inference the $IIF$ weights are injected. This technique achieves slightly worse results than decoupled $IIF$, because it does not involve re-training the classifier. However, it does not cost additional computing resources and it is useful when there are computing limitations. In particular for CIFAR datasets, post-processing $IIF$ drops the performance by $-0.9\%$ while for ImageNet-LT it achieves the same result compared to the decoupling strategy. This is due to the fact that, the CIFAR-LT datasets have larger variance than ImageNet-LT thus the decoupling strategy allows the model to explore better solutions and achieve slightly better results. 

In the end, decoupled-$IIF$ is best as it improves the performance in CIFAR10-LT by $5.5\%$, in CIFAR100-LT by $5.9$ and in ImageNet-LT by $3.8\%$ compared to Softmax. Regarding the datasets, the best performance boost is observed in CIFAR100-LT, because this dataset has larger vocabulary than CIFAR10-LT and it is less complex than ImageNet-LT.
\begin{table}
    \centering
    \caption{$IIF$ strategies on long-tailed datasets}
    \begin{tabular}{c|c|c|c|c}
    \hline
         Method&Strategy&Cifar10-LT&Cifar100-LT&ImageNet-LT  \\
         \hline
         Softmax&\multirow{2}{*}{End-to-End}&78.6&43.0&52.2\\
         $IIF_{CSL}$&&79.7&40.1&52.0\\
         \hline
         \multirow{2}{*}{$IIF$}&Post-Process&83.2&48.0&\textbf{56.0}\\
         &Decoupled&\textbf{84.1}&\textbf{48.9}&\textbf{56.0} \\
         \hline
    \end{tabular}
    
    \label{tab:img_int_strat}
\end{table}

\begin{table}
    \centering
    \caption{Post-processing $IIF$ variants on long-tailed datasets}
    \begin{tabular}{c|c|c|c}
    \hline
         Variant&CIFAR10-LT&CIFAR100-LT&ImageNet-LT  \\
         Softmax&78.6&43.0&52.2\\
         \hline
         Raw/Base2/Base10&83.2&48.0&\textbf{56.0}\\
         Smooth&\textbf{84.0}&\textbf{48.3}&55.9\\
         Rel&77.2&47.9&\textbf{56.0}\\
         Gombit&81.2&48.0&\textbf{56.0}\\
         Normit&77.4&48.2&55.3\\
         \hline
    \end{tabular}
    \label{tab:post_iif_classif_variants}
\end{table}

\begin{table}
    \centering
    \caption{$IIF$ variants with decoupled strategy and random sampling}
    \begin{tabular}{c|c|c|c}
    \hline
         Variant&CIFAR10-LT&CIFAR100-LT&ImageNet-LT  \\
         Softmax&78.6&43.0&52.2\\
         \hline
         Raw&84.1&48.9&\textbf{56.0}\\
         Smooth&\textbf{84.6}&48.8&55.8\\
         Rel&81.2&48.8&\textbf{56.0}\\
         Base2&84.1&48.9&55.9\\
         Base10&84.4&48.9&\textbf{56.0}\\
         Gombit&82.9&\textbf{49.0}&55.9\\
         Normit&80.5&48.6&55.1\\
    \hline
    \end{tabular}
    \label{tab:decoup_r_iif_classif_variants}
\end{table}

\subsubsection{$IIF$ Variants}
Next, we explore the $IIF$ variants listed in Table \ref{tab:link_functions} with respect to the post-processing strategy and the decoupled training strategy. 
Starting from the post-processing $IIF$ strategy, as Table \ref{tab:post_iif_classif_variants} indicates, the best variant for CIFAR10-LT and CIFAR100-LT datasets is smooth $IIF$ that improves the performance by $5.4\%$ and $5.3\%$ respectively. 
For ImageNet-LT the best variants are the raw, gombit and relative $IIF$ as they boost performance by $3.8\%$. 
Other variants produce similar results for ImageNet-LT, except for Normit $IIF$. This is because ImageNet-LT has a large vocabulary and the majority of its class probabilities are within a specific range of values that cause similar re-weighting for most $IIF$ variants.

In the end, smooth $IIF$ is the best choice as it generalises better than other variants and achieves the best performance in both small and large vocabulary datasets under various imbalance factors. 

Notice that the variants raw, base2 and base10 have the same performance (i.e. $83.2\%$, $48.0\%$, $56.0\%$ for CIFAR10-LT, CIFAR100-LT and ImageNet-LT respectively) under the post-processing strategy. That's because they produce exactly the same rankings, however, when using the decoupled training strategy, they have slightly different results due to different optimisation.

To illustrate this, we use the decoupled $IIF$ strategy with random sampling. As Table \ref{tab:decoup_r_iif_classif_variants} suggests, smooth $IIF$ has the best performance for CIFAR10-LT as it boosts the performance by $6.0\%$. Regarding CIFAR100-LT, the gombit $IIF$ has the best performance as it surpasses Softmax by $6.0\%$. Finally, in ImageNet-LT the raw, the relative and the base10 have the best performances boosting the accuracy by $3.8\%$.

Under the decoupling strategy, we notice that for datasets CIFAR100-LT and ImageNet-LT all variants except for normit $IIF$ produce similar results and their differences in performance are marginal. This is because the class probabilities in these datasets are within a small range of values that produce similar weights when using the aforementioned $IIF$ variants. In the end, the smooth $IIF$ is the best variant as it achieves the best performance in CIFAR10-LT and generalises well to both CIFAR100-LT and ImageNet-LT.

In conclusion, we use decoupled strategy as it produces better results than the post-processing $IIF$ strategy. Regarding the variants, we use smooth $IIF$ because it provides good performance and it generalises better than other $IIF$ variants in all datasets and strategies.

\subsection{Comparison with other Methods}
Long-tailed image classification has been advancing rapidly during the recent years and diverse solutions have been proposed. We compare our method against many families of methods such as: 
\begin{itemize}
    \item \textbf{Two Stage Methods.} We show the efficacy of our $IIF$ by comparing it to other two stage methods such as DisAlign \cite{zhang2021distribution}, LWS \cite{Kang2020Decoupling}, cRT \cite{Kang2020Decoupling} and MiSLAS \cite{zhong2021improving}.
    \item \textbf{Self-supervised.} We highlight the simplicity and stronger performance of $IIF$ against self-supervised methods such as Hybrid SC \cite{wang2021contrastive} and DRO-LT \cite{samuel2021distributional}.
    \item \textbf{Higher Capacity Models.} $IIF$ is additionally compared against higher capacity models like ensemble RIDE \cite{wang2020long}, knowledge distilled CBD \cite{iscen2021cbd} and DiVE \cite{He_2021_ICCV} and two branch network BBN \cite{zhou2020bbn}.
    \item \textbf{Margin Adjustment.} Finally, $IIF$ is compared with other margin adjustment techniques like Balanced Softmax \cite{NEURIPS2020_2ba61cc3}, LADE \cite{hong2021disentangling} and Logit Adjustment \cite{menon2021longtail}.
\end{itemize}

In summary, for our models we use smooth $IIF$ with decoupled strategy as this produces the best performance. We compare $IIF$ in common long-tailed classification benchmarks such as CIFAR-LT, ImageNet-LT and Places-LT and we show that $IIF$ surpasses the state-of-the-art.

\subsubsection{ImageNet-LT}
Our method has on average better top-1 accuracy than all state-of-the-art methods on ImageNet-LT as shown in Table \ref{tab:imagent_sota_comp}.
Our $IIF$ significantly surpasses the best two-stage DisAlign method by $2.9\%$ on average accuracy using ResNet50 and by $2.8\%$ using ResNeXt50. Secondly, it overcomes the best margin adjustment LADE method by $3.2\%$ using ResNeXt50 under a similar training budget. Additionally, it outperforms higher capacity models like ensemble RIDE by $1.4\%$  and self-supervised models like DRO-LT by $2.3\%$ using ResNet50. Furthermore, it outperforms knowledge distilled models like CBD by $4.2\%$ using ResNet50 and DiVE by $3.1\%$ using ResNeXt50, having a more straightforward training pipeline.

\begin{table}
    \centering
    \caption{Comparative Results on ImageNet-LT test set}
    \begin{tabular}{c|c|c}
    \hline
    \multirow{1}{*}{Method}&\multicolumn{1}{c|}{ResNet50}&\multicolumn{1}{c}{ResNeXt50}\\
    \hline
    Softmax&52.2&52.8\\
    \hline
    cRT\cite{Kang2020Decoupling}&47.3&49.6\\
    LWS\cite{Kang2020Decoupling}&47.7&49.9\\
    Logit Adjustment Loss\cite{menon2021longtail}&51.1&-\\
    Logit Adjustment Post-Hoc\cite{menon2021longtail}&50.3&-\\
    TDE\cite{tang2020long}&-&51.8\\
    EQL\cite{tan2020equalization}&-&46.0\\
    Seesaw\cite{wang2021seesaw}&-&50.4\\
    CBD \cite{iscen2021cbd}&51.6&-\\
    LADE \cite{hong2021disentangling}&-&53.0\\
    NorCal \cite{pan2021model}&49.7&-\\
    MisLAS \cite{zhong2021improving}&52.7&-\\
    DiVE \cite{He_2021_ICCV}&-&53.1\\
    DRO-LT \cite{samuel2021distributional}&53.5&-\\
    DisAlign\cite{zhang2021distribution}&52.9&53.4\\
    RIDE (2 experts) \cite{wang2020long}&54.4&55.9\\
    \hline
    $IIF$ (ours)&\textbf{55.8}&\textbf{56.2}\\
    \hline
    \end{tabular}
    \label{tab:imagent_sota_comp}
\end{table}

\subsubsection{CIFAR-LT}
Our method shows great performance on the CIFAR-LT datasets as well, highlighting its generalisation ability. As Table \ref{tab:img_cls_sota_cifar} suggests, $IIF$ surpasses the best margin adjustment method LADE \cite{hong2021disentangling} by $3.4\%$ on CIFAR100-LT. Moreover, it overcomes the best two-stage MisLAS by $2.5\%$ on CIFAR10-LT and by $1.8\%$ on CIFAR100-LT. Furthermore, it outperforms self-supervised methods like the Hybrid SC method \cite{wang2021contrastive} by $3.2\%$ on CIFAR10-LT and by $2.1\%$ on CIFAR100-LT. Finally it is better than ensemble methods like RIDE by $1.8\%$ on CIFAR100-LT, using a single model.

\subsubsection{Places-LT}
Finally, in Table \ref{tab:places_sota} the results on Places-LT are displayed. $IIF$ outperforms all other methods on average accuracy, achieving $40.2\%$ top-1 accuracy. It achieves an absolute $9.1\%$ increase compared to Softmax and $4.3\%$ increase compared to OLTR \cite{liu2019large}. Additionally, it surpasses the margin adjustment LADE method, by an overall $1.4\%$ in top-1 accuracy and the two-stage DisAlign by $0.9\%$.

\begin{table}[t]
    \centering
    \caption{Results on CIFAR-LT datasets using imbalance ratio 100}
    \begin{tabular}{c|c|c}
    \hline
         Dataset&\multicolumn{1}{c|}{CIFAR10-LT}&\multicolumn{1}{c}{CIFAR100-LT}\\
         \hline
         Softmax&78.6&43.0\\
         \hline
         Logit-Adjustment PostHoc \cite{menon2021longtail}&78.9&43.2\\
         Logit-Adjustment Loss \cite{menon2021longtail}&79.1&43.0\\
         LDAM-DRW \cite{cao2019learning}&77.0&42.0\\
         LDAM-DRW-RSG\cite{wang2021rsg}&79.6&44.6\\
         BBN\cite{zhou2020bbn}&79.2&42.6\\
         CBD \cite{iscen2021cbd}&-&44.8\\
         DiVE \cite{He_2021_ICCV}&-&45.4\\
         TailCalibX + CBD \cite{vigneswaran2021feature}&-&46.6\\
         NorCal \cite{pan2021model}&77.8&-\\
         LADE \cite{hong2021disentangling}&-&45.4\\
         DRO-LT \cite{samuel2021distributional}&-&47.3\\
         Hybrid SC \cite{wang2021contrastive}&81.4&46.7\\
         Hybrid SPC \cite{wang2021contrastive}&78.8&45.0\\
         MiSLAS \cite{zhong2021improving}&82.1&47.0\\
         RIDE (2 experts) \cite{wang2020long}&-&47.0\\
         \hline
         $IIF$ (ours)&\textbf{84.6}&\textbf{48.8}\\
         \hline
    \end{tabular}
    \label{tab:img_cls_sota_cifar}
\end{table}

\begin{table}[htb]
    \centering
    \caption{Results on Places-LT}
    \begin{tabular}{c|c}
    \hline
         Method&Top-1 Accuracy \\
         \hline
         Softmax&31.1\\
         \hline
         OLTR \cite{liu2019large}&35.9\\
         LWS \cite{Kang2020Decoupling}&37.6\\
         cRT \cite{Kang2020Decoupling}&36.7\\
         Balanced Softmax \cite{NEURIPS2020_2ba61cc3}&38.7\\
         DisAlign\cite{zhang2021distribution}&39.3\\
         LADE \cite{hong2021disentangling}&38.8\\
         \hline
         $IIF$ (ours) &\textbf{40.2} \\
    \hline
    \end{tabular}
    
    \label{tab:places_sota}
\end{table}

\subsubsection{Comparison against LWS}
Our $IIF$ significantly surpasses the LWS method in both ImageNet-LT and Places-LT datasets. 
LWS \cite{Kang2020Decoupling} uses multiplicative adjustment as shown in Table \ref{tab:sim_adj_methods}, like our $IIF$. However, the class margins in LWS, need to be learned in two stage training, whereas in $IIF$, the margins can be injected during inference using Post-Process $IIF$, without the need for classifier retraining. Furthermore, the margins of $IIF$ are easier to explain as they are calculated directly from the training dataset, whereas the LWS margins are learnable and more difficult to explain.

\section{Long-tailed Instance Segmentation}
In the previous section, we showed that $IIF$ can achieve good performance in long-tailed classification. In this section, we show that $IIF$ can generalise to downstream tasks such as long-tailed instance segmentation. 

\subsection{Experiment Setup}
\subsubsection{Dataset}
We use LVIS version 1 (LVISv1) which contains 99k images for training and 19.8k images for validation. LVISv1 is a heavily class imbalanced dataset that contains $1,203$ categories that are grouped according to their image frequency into \textit{rare} categories (those with 1-10 images in the dataset), \textit{common} categories (11 to 100 images in the dataset) and \textit{frequent} categories (those with $>100$ images in the dataset). 
We report our results using average mask precision $AP$, average mask precision for rare $AP_r$, common $AP_c$ and frequent categories $AP_f$ and average box precision $AP_b$.
The imbalance factor of LVISv1 is shown in Table \ref{tab:imb_factor}.

\subsubsection{Implementation Details}
We use a plethora of architectures such as MaskRCNN \cite{he2017mask}, Cascade MaskRCNN \cite{cai2019cascade} and Hybrid Task Cascade \cite{chen2019hybrid}. For our intermediate experiments, we use the 1x schedule in order to reduce the computational time and still showcase the performance of $IIF$. When we compare against the state-of-the-art we use a longer training schedule that is 2x and standard model enhancements such as Cosine Classifier\cite{zhang2021distribution} and Normalised Mask\cite{wang2021seesaw}. We also use RFS \cite{gupta2019lvis} as our sampling policy and FASA \cite{zang2021fasa} as our augmentation policy and we train all models using the MMdetection framework \cite{chen2019mmdetection}.

\subsection{$IIF$ in Long-tailed Instance Segmentation}
We analyse $IIF$ in conjunction to training strategies, sampling strategies, $IIF$ variants and model architectures.
Unless specified, for all experiments we use MaskRCNN with ResNet50 as our main architecture.

\subsubsection{Training Strategies}
The task of long-tailed instance segmentation is different and more complex than long-tailed classification as it contains the special background class, it has a larger imbalance factor as shown in Table \ref{tab:imb_factor} and the target distribution is not uniform but long-tailed.

For this reason, we examine two strategies of applying $IIF$: either end-to-end training or decoupled strategy. As shown in Table \ref{tab:e2e_vs_decoup} the best strategy is end-to-end training, as this gives the best mask $AP$ and box $AP_b$. In detail, the 12-epoch schedule $IIF$ boosts mask $AP$ by $4.6\%$ and box $AP$ by $2.6\%$ while the 24-epoch schedule increases mask $AP$  by $3.4\%$ and box $AP$ $2.5\%$ compared to Softmax.
The decoupled strategy also increases the performance by $3.1\%$ in mask $AP$ and by $1.9\%$ in box $AP$ compared to Softmax trained for 12-epochs. However, decoupling strategy costs more training resources and achieves lower performance than the end-to-end training. To this end, end-to-end trainning is better for long-tailed instance segmentation in contrast to long-tailed classification where decoupled training works best. The reason is that the long-tail instance segmentation task is a finetuning task which typically uses a backbone pretrained on ImageNet-1K. Thus, the network has already learned good representations \cite{Kang2020Decoupling} and $IIF$ can be used end-to-end to finetune the model in the downstream task. Also, end-to-end training is preferable because it converges faster than decoupled training as shown in Table \ref{tab:e2e_vs_decoup}.

Finally, end-to-end $IIF$ training is a superior because it reduces not only the foreground imbalance but also the foreground to background imbalance, allowing the model to distinguish rare categories from the background.

\begin{table}
    \centering
    \caption{End-to-end against decoupled strategy using $IIF$}
    \begin{tabular}{c|c|cc}
    \hline
        \multirow{2}{*}{MaskRCNN}&\multirow{2}{*}{Epochs}& \multicolumn{2}{c}{LVISv1}  \\
        & &$AP_{b}$ &$AP$ \\
        \hline
         Softmax&12&16.9& 15.2 \\
         Softmax&24&19.5& 18.7 \\
        \hline
         End to End&12&19.5& 19.8 \\
         End to End&24&\textbf{22.0}&\textbf{22.1} \\
         Decoupled&12/12&18.8& 18.3\\
    \hline
    \end{tabular}
    \label{tab:e2e_vs_decoup}
\end{table}

\begin{table}
    \centering
    \caption{Sampling strategies using $IIF$}
    \begin{tabular}{c|c|c|cccccc}
    \hline
         E2E&Epochs&Sampler&$AP$ &$AP_r$  &$AP_c$  &$AP_f$&$AP_{b}$  \\
         \hline
         \multirow{2}{*}{\checkmark}&12&\multirow{2}{*}{rand}&19.8&7.5&17.7&27.6&19.5\\
         &24&&22.1&9.9&20.3&\textbf{29.6}&22.0\\
         \hline
         \multirow{4}{*}{\checkmark}&12&\multirow{4}{*}{RFS}&22.6&\textbf{12.6}&21.7&28.0&22.6\\
         &16&&22.8&11.9&21.8&28.7&23.0\\
         &18&&22.8&11.9&21.7&28.9&23.1\\
         &24&&\textbf{22.9}&10.9&\textbf{21.9}&29.2&\textbf{23.5}\\
         \hline
         &12/12&rand/rand&18.3&5.4&15.9&26.7&18.8\\
         &12/12&rand/RFS&19.0&8.0&16.5&26.6&19.2\\
         &12/12&RFS/RFS&18.8&6.9&16.6&26.5&19.5\\
    \hline
    \end{tabular}
    \label{tab:iif_rfs_ltis}
\end{table}

\begin{table}
    \centering
    \caption{Comparative segmentation results on (M)askRCNN\cite{he2017mask}, (C)ascade Mask-RCNN \cite{cai2019cascade} and (H)ybrid Task Cascade \cite{chen2019hybrid} using (R)esnet \cite{he2016deep} or Resne(X)t\cite{xie2017aggregated}}
    \begin{tabular}{c|c|p{0.5cm}p{0.5cm}p{0.5cm}p{0.5cm}p{0.5cm}p{0.5cm}}
    \hline
         Method& Framework& $AP$&$AP_{50}$&$AP_{75}$ &$AP_r$  &$AP_c$  &$AP_f$ \\
         \hline
         Softmax&\multirow{2}{*}{M.R50} &15.2 &24.4&16.1& 0.0& 10.6 &26.9\\
         IIF& &\textbf{19.8} &\textbf{32.3}&\textbf{20.7}&\textbf{7.5}&\textbf{17.7}&\textbf{27.6}\\
         \hline
         Softmax&\multirow{2}{*}{M.R101} &16.7&26.5&17.6 & 0.5& 12.5 &28.5\\
         IIF& & \textbf{21.3}&\textbf{34.3}&\textbf{22.1}&\textbf{7.5}&\textbf{19.5}&\textbf{29.2}\\
         \hline
         Softmax&\multirow{ 2}{*}{M.X101} &18.6&29.1&19.6 &0.6&14.5 &31.1\\
         IIF& & \textbf{23.5}&\textbf{37.4}&\textbf{24.9}&\textbf{9.2}&\textbf{21.9}&\textbf{31.5}\\
         \hline
         Sofmax&\multirow{ 2}{*}{C.R101}  &18.8&28.7&20.1  &0.6 &15.7 &30.3 \\
         IIF& & \textbf{24.2}&\textbf{36.6}&\textbf{25.8}&\textbf{9.5}&\textbf{23.8}&\textbf{31.0}\\
         \hline
         Sofmax&\multirow{ 2}{*}{H.R101}  &19.1&28.9&20.5  &0.6  &15.8  &31.0 \\
         IIF& &\textbf{24.7}&\textbf{36.9}&\textbf{26.5}&\textbf{9.3} &\textbf{24.4}&\textbf{31.9}\\
    \hline    
    \end{tabular}
    \label{tab:lvis_instance_segm}
\end{table}

\subsubsection{Sampling Strategies}
Next, we examine sampling strategies, in particular, oversampling and random sampling. In contrast to long-tailed classification, the oversampling strategy is essential to the performance of long-tailed instance segmentation methods and many works use it \cite{NEURIPS2020_2ba61cc3,wang2021seesaw,hsieh2021droploss,pan2021model,Feng_2021_ICCV}. Similar to these works, we explore RFS sampling \cite{gupta2019lvis}, and random sampling and we compare them with end-to-end training and decoupling strategy.
As shown in Table \ref{tab:iif_rfs_ltis}, the best sampling strategy is RFS \cite{gupta2019lvis} used in End-to-End (E2E) for 24 epochs as this has the best overall $AP$. In detail, it achieves $22.9\%$ in overall mask $AP$ and  $23.5\%$ in overall box $AP$. It also increases the $AP_r$ by $1.0\%$ and $AP_c$ by $1.6\%$ compared to random sampling used in End-to-End training for 24 epochs. However, this technique reduces the performance of frequent categories slightly by $0.4\%$ compared to end-to-end random sampling, as also noted by \cite{gupta2019lvis}. Moreover, the end-to-end 12-epoch RFS schedule achieves the best $AP_r$ adding a further boost of $1.7\%$, but at the same time it lowers the performance of the frequent categories $1.6\%$, compared to the end-to-end 24-epoch RFS schedule. This indicates that there exists a trade-off for frequent and rare categories that depends on the training schedule, i.e. the longer schedule may be suboptimal for the rare categories but it benefits frequent categories and vice versa. Regarding the decoupling strategy, we use three different sampling combinations for the two stage training; random sampling for both stages, random sampling first and RFS secondly and finally RFS for both stages. All decoupling strategies require more training resources and have worst performance than training end-to-end. This is because, in long-tailed instance segmentation, the backbone is already pretrained on Imagenet-1K, thus the decoupled strategy converges slower than the end-to-end training. In the end, we adopt the end-to-end 24-epoch schedule as this has the best overall performance.

\subsubsection{Extension to Deeper Architectures}
We show the generalisability of $IIF$ by applying it to the popular instance segmentation models such as MaskRCNN, Cascade MaskRCNN and Hybrid Task Cascade using 1x schedule and random sampling. As shown in Table \ref{tab:lvis_instance_segm}, $IIF$ improves the performance of all models significantly. Furthermore, the gains in performance become larger as models become deeper linearly, which indicates that our method can generalise well to larger architectures. $IIF$ increases MaskRCNN ResNet50 by $4.6\%$, MaskRCNN ResNet101 by $4.6\%$, MaskRCNN ResNeXt101 by $4.9\%$, Cascade MaskRCNN ResNet101 by $5.4\%$ and Hybrid Task Cascade ResNet101 by $5.6\%$ in overall mask $AP$. Moreover, $IIF$ increases the performance of all categories, both head and tail, for all architectures. This is due to the fact that even frequent categories may have lower expectations compared to the dominant background class, especially for the edge locations of an image. $IIF$ can alleviate such imbalance and increase the performance for all categories, thus it is a robust method for long-tailed instance segmentation.

\begin{table*}[t]
    \centering
    \caption{Ablation study of $IIF$ variants with MaskRCNN on LVIS}
    
    \begin{tabular}{c|c|cccccc|cccccc}
\hline
\multicolumn{2}{c|}{LVISv1.0}&\multicolumn{6}{c|}{Box AP}&\multicolumn{6}{c}{Mask AP}\\
\hline
Variant& Method & $AP$ & $AP_{50}$ & $AP_{75}$ & $AP_r$ & $AP_c$ & $AP_f$& $AP$ & $AP_{50}$ & $AP_{75}$ & $AP_r$ & $AP_c$ & $AP_f$\\
\hline
Baseline& Softmax &16.1 &26.5  & 16.9& 0.4 & 10.5&29.2&15.2 &24.4  & 16.1& 0.5 & 10.6&26.9\\
\hline
Raw&\multirow{7}{*}{$IIF$}&19.5 &34.7 &18.5 &6.6 &15.4 &29.7&19.8 &32.3 &20.7 &7.5 &17.7 &27.6\\
Smooth& &19.0 &34.3  &18.0 &5.4 &14.9 &29.6&19.5 &32.0  &20.3 &6.8 &17.3&27.6\\
Relative& &19.7 &\textbf{34.8}  &18.9 &\textbf{6.8} &15.5 &30.0&\textbf{19.9} &\textbf{32.4} &20.8 &7.3 &17.8 &\textbf{27.9}\\
Base2&  &19.0 &34.3  & 17.7& 6.4 & 14.5&29.5&19.5 &31.9  & 20.4&7.6 & 17.0&27.6\\
Base10&  &19.5 &33.2  & 19.7& 3.2 & 16.6&29.9&19.2 &30.9  & 20.2& 4.2 & 17.7&27.4\\
Normit&  &19.3 &33.1  & 19.5& 2.5 & 16.3&\textbf{30.1}&19.0 &30.7  & 19.9& 3.7 & 17.2&27.7\\
Gombit&  &19.7 &34.7  & 19.1&5.9& 16.0&29.9&19.8 &32.3  & 20.7&6.9& 17.9&27.7\\
\hline
Raw&\multirow{7}{*}{$IOF$}&19.0 &34.2 &17.7 &6.0 &14.6 &29.6&19.5 &32.1 &20.3 &7.4 &17.1 &27.6\\
Smooth& &18.7 &34.0  &17.6 &5.8 &14.4 &29.3&19.5 &31.8  &20.2&7.4 &17.0 &27.5\\
Relative& &19.0 &34.1  &17.9 &6.6 &14.4 &29.6&19.5 &31.9  &20.2 &\textbf{7.8} &17.0 &27.5\\
Base2&  &18.2 &33.8  & 16.6& 5.6 & 13.6&28.9&19.1 &31.6  & 19.9& 7.1 & 16.4&27.3\\
Base10&  &\textbf{20.0} &34.2  & \textbf{20.3}& 4.9 & \textbf{17.0}&30.0&\textbf{19.9} &32.0  & \textbf{20.9}& 6.2 & \textbf{18.2}&27.7\\
Normit&  &19.4 &33.3  & 19.4& 2.9 & 16.3&\textbf{30.1}&19.1 &30.8  & 20.0& 3.7 & 17.4&27.7\\
Gombit&  &18.9 &34.3  & 17.4&5.0& 14.7&29.6&19.5 &32.1 & 20.3&7.0& 17.2&27.5\\
\hline
\end{tabular}
\label{tab:abl_iif_lvis}
\end{table*}

\begin{table}
    \centering
    \caption{Ablation study of components used with $IIF$}
    \begin{tabular}{ccccc|c}
    \hline
         \rotatebox[origin=c]{0}{Cos. Cls.}&
         \rotatebox[origin=c]{0}{Norm. M.}&
         \rotatebox[origin=c]{0}{FASA}&
         \rotatebox[origin=c]{0}{RFS}&base10-$IOF$&
         $AP$\\
         \hline
         &&&&&18.7\\
         \checkmark&&&&&20.1\\
         \checkmark&\checkmark&&&& 20.8\\
        \checkmark&\checkmark&\checkmark&&&23.3\\
        \checkmark&\checkmark&\checkmark&\checkmark&&25.0\\
        \hline
        \checkmark&\checkmark&\checkmark&&\checkmark&24.1\\
        \checkmark&\checkmark&\checkmark&\checkmark&\checkmark&26.3\\
        \hline
    \end{tabular}
    \label{tab:iif_ltis_enhancements}
\end{table}

\begin{table*}[htb]
    \centering
        \caption{Comparison against the state-of-the-art on LVISv1.0, using MaskRCNN. The symbol $^\dagger$ denotes that the results have been reproduced.}
        \begin{tabular}{c|c|c|c|c|c|c}
        \hline
        Method&Backbone &$AP$&$AP_r$&$AP_c$&$AP_f$&$AP_b$ \\
         \hline
          Softmax&\multirow{ 10}{*}{ResNet-50-FPN}&18.7&1.1&16.2&29.2&19.5\\
          EQL\cite{tan2020equalization}&&21.6&3.8&21.7&29.2&22.5 \\
          DropLoss\cite{hsieh2021droploss}&&22.3&12.4&22.3&26.5&22.9 \\
          
          Forest-RCNN\cite{wu2020forest}&&23.2&14.2&22.7&27.7&24.6 \\
          RFS$^\dagger$\cite{gupta2019lvis}&&23.7&13.3&23.0&29.0&24.7\\
          FASA \cite{zang2021fasa} &&24.1&17.3&22.9&28.5&-\\
          DisAlign \cite{zhang2021distribution} & &24.2&13.2&23.8&29.3& 24.7\\
          NorCal \cite{pan2021model} &&25.2&\textbf{19.3}&24.2&29.0&\textbf{26.1}\\
          EQLv2\cite{tan2021eqlv2}& &25.5&17.7&24.3&30.2& 26.1\\          
          $IIF$ (ours) & &\textbf{26.3}&18.6&\textbf{25.2}&\textbf{30.8}&25.8\\
          \hline
          Softmax&\multirow{ 7}{*}{ResNet-50-FPN (RSB)}&23.4&8.4&22.5&30.8&23.1\\
          EQL$^\dagger$\cite{tan2020equalization}&&23.9&14.0&23.4&28.9&236 \\
          RFS$^\dagger$\cite{gupta2019lvis}&&25.4&13.0&25.5&30.9&24.9\\
          FASA$^\dagger$ \cite{zang2021fasa}&&25.5&14.3&25.2&30.7&24.9\\
          DropLoss$^\dagger$\cite{hsieh2021droploss}&&25.7&14.4&26.6&29.7&25.1 \\
          NorCal$^\dagger$ \cite{pan2021model}&&27.1&18.4&26.6&31.5&26.8\\
          $IIF$ (ours)&&\textbf{27.4}&\textbf{19.4}&\textbf{26.8}&\textbf{31.5}&\textbf{27.4}\\
        \hline
    \end{tabular}
    \label{tab:lvis1_sota}
\end{table*}

\subsubsection{$IIF$ Variants}
We conduct an extensive ablation study of different $IIF$ variants in Table \ref{tab:abl_iif_lvis}. All $IIF$ variants significantly improve the baseline in both mask and box $AP$ and the best variant is base10 $IOF$.

The base10 $IOF$ achieves $20.0\%$ overall box $AP$ and $19.9\%$ overall mask $AP$ and it boosts the detection performance by $4.5\%$ and segmentation performance by $5.7\%$ for rare categories. There are other variants that achieve better segmentation performance for rare categories like the relative $IIF$ that boosts performance by $7.3\%$. However, in the task of long-tailed instance segmentation it is better to opt for a variant that achieves high bounding box performance, as this enables the mask $AP$ to improve further by combining this technique with other sampling strategies and methods. In our experiments we have observed that, during MaskRCNN inference, the bounding box performance, determines the segmentation performance, thus the bounding box performance is the bottleneck.  For this reason, we use base10 $IOF$ to achieve the best possible box $AP$ and this enables the creation of better models as we show in the following section.

\subsubsection{$IIF$ Enhancements}
We use the base10 $IOF$ variant and end-to-end training. Moreover, we use standard techniques that have been previously used by other state-of-the-art such as Normalisation Mask \cite{wang2021seesaw}, Cosine classifier \cite{zhang2021distribution}, RFS \cite{gupta2019lvis} and FASA \cite{zang2021fasa}. Additionally, we use a stricter Non-maximum suppression threshold that is $0.3$, mask threshold of $0.4$ and a longer training schedule that is 2x. 

Starting from the Softmax model, we replace the Dot-product Classifier wit Cosine Classifier following \cite{zhang2021distribution}, this adds $1.4\%$ in mask $AP$. Next, we adopt a Normalisation Mask \cite{wang2021seesaw} that further increases the performance by $0.7\%$. Recently, Zang et. al. proposed FASA \cite{zang2021fasa} which is a novel feature augmentation technique. Using FASA we further improve the model by $2.5\%$. Using $IIF$ in addition to these methods, the model's performance is further increased by $0.8\%$.
If we adopt RFS \cite{gupta2019lvis} as our sampling strategy, this further increments the performance by $1.7\%$ compared to FASA. Finally, using base10 $IOF$, a stricter NMS threshold of $0.3$ and mask threshold of $0.4$,  we further increase performance by $1.3\%$ achieving $26.3\%$ in overall mask $AP$.

\subsection{Comparison to Other Methods}
We compare our $IIF$ method against the state-of-the-art in Table \ref{tab:lvis1_sota}. Using ResNet50, our method has the best overall segmentation performance, surpassing EQLv2 \cite{tan2021eqlv2} by $0.8\%$ and NorCal \cite{pan2021model} by $1.1\%$ in overall $AP$. 
Furthermore, our method achieves the best AP in common and frequent categories. Also, it increases the $AP$ by $7.6\%$, $AP_r$ by $17.5\%$, $AP_c$ by $9.0\%$, $AP_f$ by $1.6\%$ and $AP_b$ by $6.3\%$ compared to vanilla Softmax.

We further investigate the performance of ResNet50-RSB \cite{wightman2021resnet} which is a ResNet50 backbone pretrained with better augmentations and regularisations. As the compared methods did not use this backbone, we have reproduced them for fair comparison.
Using ResNet50-RSB \cite{wightman2021resnet}, $IIF$ surpasses the state-of-the-art in overall mask and box performance reaching $27.4\%$ in both metrics. Also, it  outperforms NorCal \cite{pan2021model} by $0.3\%$ and RFS \cite{gupta2019lvis} by $1.7\%$. Moreover, it achieves the best performance in rare, common and frequent categories reaching $19.4\%$, $26.9\%$ and $32.1\%$ respectively.

We notice that $IIF$ generally improves all categories, which is different from long-tailed image classification where there is a performance trade-off between rare and frequent categories. This is because in long-tailed instance segmentation the trade-off is not only between foreground but also between background and foreground categories. In long-tailed segmentation, the background samples dominate the training process and render all foreground classes as the minority. Thus, using $IIF$ all categories can benefit resulting in the general performance boost. 

\subsection{Model Analysis}
\begin{figure*}
    \centering
    \includegraphics[scale=0.19]{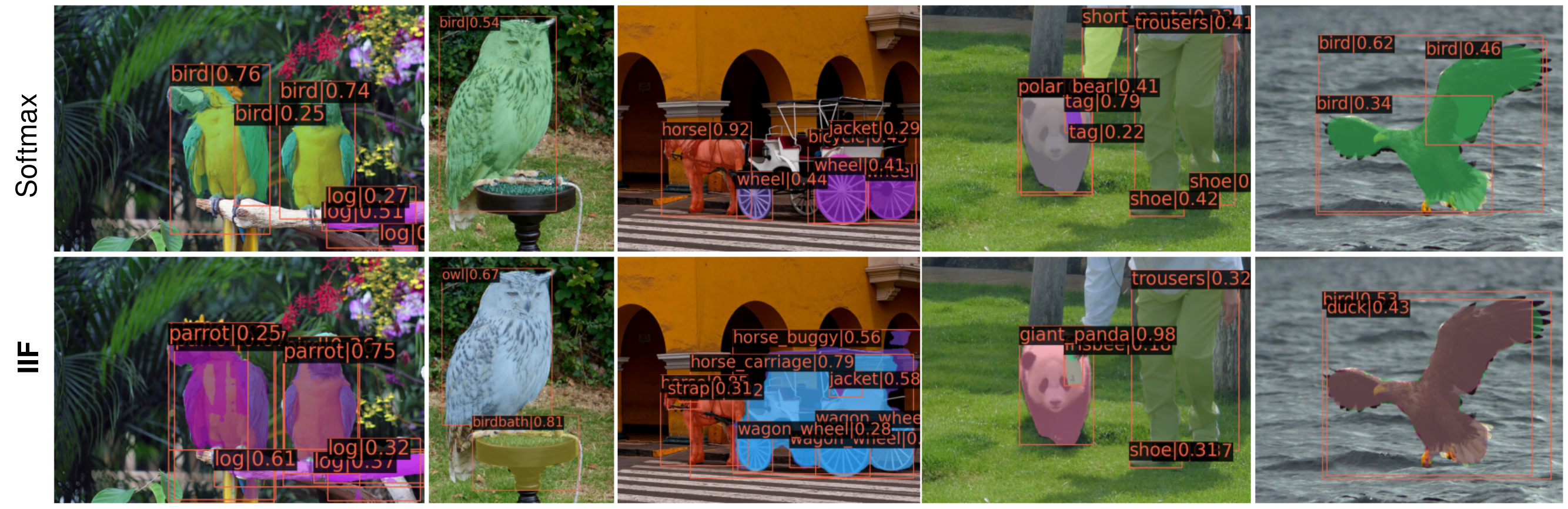}
    \caption{MaskRCNN-ResNet50 detections on LVISv1 validation set using Softmax versus our $IIF$ method. $IIF$ can correctly detect rare classes such as \textit{parrot, owl, horse-carriage} and \textit{giant panda} in contrast to Softmax method. However, both methods fail to detect the rare class \textit{eagle} in the last image.}
    \label{fig:maskrcnn_det}
\end{figure*}

\begin{figure}
    \centering
    \includegraphics[scale=0.46]{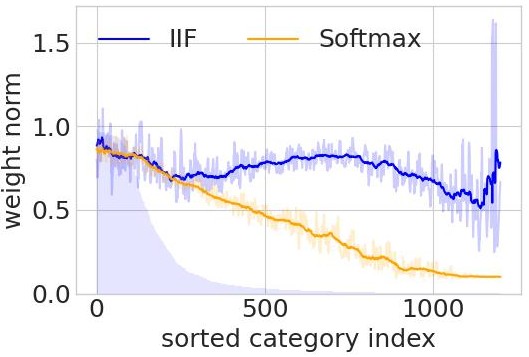}
    \caption{Visualisation of MaskRCNN classifier's weight norms on LVIS using Softmax and $IIF$. $IIF$ produces a more balanced weight-norm distribution in comparison to Softmax, thus it reduces the frequent category bias.}
    \label{fig:iif_weight_norms}
\end{figure}

Inspired by \cite{Kang2020Decoupling} we analyse the weight norms of the classification layer of MaskRCNN trained with our method. As seen in Figure \ref{fig:iif_weight_norms}, $IIF$ produces a more balanced weight norm distribution compared to Softmax. In this way, it removes classification bias by increasing the norms associated with rare classes and decreasing the norms associated with frequent classes.
Lastly, we compare instance segmentation results of our method against Softmax, shown in five images from LVIS validation set using MaskRCNN in Figure \ref{fig:maskrcnn_det}. Our $IIF$ model recognises correctly the rare classes like the \textit{parrot}, \textit{owl}, \textit{horse-carriage} and \textit{giant panda}, in contrast to vanilla Softmax, that either classifies them as the common classes \textit{bird}, \textit{polar bear} or does not recognise them at all. However, not all rare categories can be correctly recognised with $IIF$ as our method did not detect the \textit{eagle} in the last image. Nevertheless, $IIF$ shows promising results as it predicted a more interesting and rare class that is \textit{duck} instead of the common class \textit{bird}, probably because the context around the image is water. This shows that this method can be further improved by explicitly modeling the context around the images or by capturing the relationship between objects inside the image.

\section{Discussions}
As presented in the above experiments, $IIF$ has proven to be a robust method that can be used in many long-tailed tasks such as long-tailed classification and long-tailed instance segmentation to boost the performance of rare categories.

Moreover, it generalises well to many backbones and architectures and therefore it can be a valuable component to long-tailed methods.

As shown in classification, $IIF$ can be used either as a post-processing strategy or as decoupled strategy. The decoupling strategy has slightly better performance than the post-processing strategy but it costs additional training. After exploring many $IIF$ variants, we showed that $IIF$ is robust and we used the smooth $IIF$ variant as this produced the best performance.
On the other hand, $IIF$ uses weight multiplication. This may be disadvantageous as it makes the model's output non-smooth and close to one-hot distribution thus it may increase the expected calibration error of the model. However, the multiplicative adjustment generalises better in downstream tasks as it produces fewer false positives than additive margin adjustment methods. 
To this end, we developed $IIF$ for long-tailed instance segmentation.
We compared the end-to-end strategy against decoupled strategy and found that the former is better. This is because end-to-end training allows the model to compensate for the background-to-foreground imbalance and foreground-to-foreground imbalance simultaneously during optimisation in a stronger fashion than decoupled strategy.

Moreover, we used $IIF$ along with sampling strategies and long-tailed techniques and we found that $IIF$ can boost their performance. By combining $IIF$ with standard enhancements, we outperformed all the state-of-the-art methods. We also showed that our $IIF$ model generally increases the performance of both frequent and rare categories as it tackles background and foreground imbalance.

However, $IIF$ has a lot of variants and choosing the right variant is not trivial as this depends on the dataset's statistics. This has been tackled by previous works using learnable margins but these may not be suitable for safety-critical applications as they are not explainable. In contrast, our $IIF$ uses dataset-dependent margins that are easy to use and achieve great performance in long-tailed classification. At the same time, $IIF$ produces fewer false positives than previous handcrafted margin-adjustment techniques in downstream tasks, thus it is a superior choice. Finally, we have also tested $IIF$ in the general object detection benchmark COCO, using both one-stage and two stage detectors, showing promising results in the Appendix.

\section{Conclusion}
\label{sec:conclusion}
In this work, we proposed the novel Inverse Image Frequency ($IIF$) to address the long-tailed problem that is a common issue in most real-world datasets.
Our method reweights the classification logits of the deep model to improve the recognition performance of the rare classes in the dataset. 
We investigated $IIF$ with many training strategies and variations on four classification datasets, one instance segmentation dataset and one object detection dataset. We showed that decoupled smooth $IIF$ works the best in the classification task; the end-to-end base-10 $IOF$ works the best in the long-tailed instance segmentation task.
Our $IIF$ models can largely improve the rare category performance and surpass the state-of-the-art by a large margin (e.g., $\sim 3.0\%$ on ImageNet-LT compared to similar methods and $\sim 0.7\%$ on LVIS in overall performance), thus it can serve as a valuable component in the long-tailed recognition methodology. Our models can be used in a variety of applications such as autonomous vehicles, Internet of Things and medical applications where data follow a long-tailed distribution. In the future, we will expand $IIF$ to other tasks such as semantic segmentation and few-shot learning and explore optimal sampling strategies to further boost the performance of rare classes.  
\section{Acknowledgements}
This work was supported by the Engineering and Physical Sciences Research Council (EPSRC) Centre for Doctoral Training in Distributed Algorithms [EP/S023445/1]; EPSRC ViTac project [EP/T033517/2]; EPSRC GNOMON: Deep Generative Models in non-Euclidean Spaces for Computer Vision \& Graphics [EP/X011364/1]; EPSRC DEFORM: Large Scale Shape Analysis of Deformable Models of Humans [EP/S010203/1]; King's College London NMESFS PhD Studentship; the University of Liverpool and Vision4ce. It also made use of the facilities of the N8 Centre of Excellence in Computationally Intensive Research provided and funded by the N8 research partnership and EPSRC [EP/T022167/1]. 

\bibliographystyle{IEEEtran}
\bibliography{ref}

 \appendix
 
\noindent \textit{Proof of Eq. 15}. Let $w_i=\log{(\frac{1}{p(i)})}$ and $\mathcal{Y}$ be the onehot encoded vector of class $y$. Then the gradient of $IIF$ Cross Entropy can be computed as follows:
\begin{equation}
    \begin{aligned}
        \frac{\partial CE_{IIF}(q,\mathcal{Y})}{\partial z_i}
        &= -\frac{\partial}{\partial z_i}\sum_{i=1}^{C}\mathcal{Y}_i\log ( q^{IIF}_{i} ) \\
        &= -\frac{\partial}{\partial z_i} \log\frac{\exp(w_y z_y)}{\sum_{j=1}^{C}\exp(w_j z_j)} \\
        &=-\frac{\partial}{\partial z_i}(w_yz_y-\log(\sum_{j=1}^{C}\exp(w_j z_j)))\\
        &=\left\{
        	\begin{array}{ll}
        		-w_i+w_i\frac{\exp(w_i z_i)}{\sum_{j=1}^{C}\exp(w_j z_j)}\\
        		w_i\frac{\exp(w_i z_i)}{\sum_{j=1}^{C}\exp(w_j z_j)} 
        	\end{array}
        \right.\\
        \\
        &=\left\{
        	\begin{array}{ll}
        		w_i(q^{IIF}_i-1) \\
        		w_iq^{IIF}_i  
        	\end{array}
        \right.\\
        \\
        &=\left\{
        	\begin{array}{ll}
        		-\log(p(i))(q^{IIF}_i-1)  & \mbox{if } i = y \\
        		-\log(p(i))q^{IIF}_i & \mbox{otherwise } 
        	\end{array}
        \right.
    \end{aligned}
\end{equation}

\noindent \textit{Proof of} $\sum_{i}^Cq_{IIF,i}=1$.  Let $w_i=\log{(\frac{1}{p(i)})}$.
\begin{gather}
    \text{For} \; C=1,\; \sum_i^1q_{IIF,i}=\frac{\exp(z_1 w_1)}{\exp(z_1 w_1)}=1. \\
     \text{For} \;C=k \in \mathbb{N}, \; \text{assume} \; \sum_{i}^kq_{IIF,i}=1 \; \text{is true.}\\
     \text{For} \;C=k+1, \; \sum_i^{k+1}q_{IIF,i}= \sum_i^{k+1} \frac{\exp(z_iw_i)}{\sum_j^{k+1}\exp(z_jw_j)}\\
     =\frac{\sum_i^{k}\exp(z_iw_i)}{\underbrace{\exp(z_{k+1}w_{k+1})}_\textit{a}+\underbrace{\sum_j^{k}\exp(z_jw_j)}_\textit{b}}+\frac{\exp(z_{k+1}w_{k+1})}{\sum_j^{k+1}\exp(z_jw_j)}\\
     =\frac{b}{a+b}+\frac{a}{a+b}=1
\end{gather}

\section{Object Detection}
Additionally, we perform experiments on MS-COCO dataset \cite{lin2014microsoft}, which in contrast to LVIS has 80 classes. COCO is considered balanced dataset because it has plenty of diverse instances per class. However, in COCO there is still imbalance between classes as categories such as \textit{person} dominate the dataset and this results in a large imbalance factor as shown in Table 1. For COCO dataset we report only bounding box $AP$ and since it does not group the classes according to their frequency, we further report tail-k $AP$ for the most rare categories, where $k$ denotes the group size. 

\subsection{Implementation Details}
For our experiments in object detection, we used YOLOv3 \cite{redmon2018yolov3}, Faster-RCNN \cite{ren2015faster}), Mask-RCNN \cite{he2017mask}, SSD \cite{liu2016ssd} and DetectoRS \cite{qiao2021detectors}. Models without our $IIF$ method are used as the baselines. All the other settings were kept the same except that we train these models with or without $IIF$.

\subsubsection{Faster RCNN}
It was implemented in PyTorch and the backbone network was the pre-trained ResNet50-FPN. We used a learning rate of 0.02, weight decay of 0.0001, the momentum of 0.9, batch size of 16 and the 2x training schedule.

\subsubsection{MaskRCNN} 
The model was implemented using MMdetection framework \cite{chen2019mmdetection} and the default training settings using a 1x schedule. 

\subsubsection{YOLOv3} 
Bayesian Optimisation was used to determine the optimal hyper-parameters of YOLOv3 and the pre-trained Darknet53 was taken as the backbone network. Furthermore, Focal Loss~\cite{lin2017focal} was used for objectness optimisation, complete IoU~\cite{zheng2020distance} for bounding box regression and Cross-Entropy for classification. The model was trained for 70 epochs using SGD, image augmentations, momentum of 0.9, weight decay of 0.0005 and an initial learning rate of 0.002 that drops by a factor of 10 at epochs 35 and 55, batch size 32, batch normalisation and multi-scale training at 640-pixel input.

\subsubsection{SSD} The SSD with VGG16 as the backbone was used. The model was trained for 120 epochs on images of 300x300 resolution using SGD and a learning rate of 0.002 that drops by a factor of 10 at epochs 80 and 110. The model was implemented using the recommended settings from the PyTorch implementation for simplicity and reproducibility. 

\subsubsection{End-to-end Training}
When we train our $IIF$ models for object detection, we use end-to-end training as we observed that a two-stage strategy does not produce better results and it costs extra training time.

\begin{table}
    \centering
    \caption{$IIF$ variants using MaskRCNN on COCO for object detection}
    \begin{tabular}{c|c|p{0.48cm}p{0.48cm}p{0.48cm}p{0.62cm}p{0.74cm}p{0.74cm}}
\hline
Variant&Method & $AP$ & $AP_{50}$ & $AP_{75}$ & tail-5 & tail-10 & tail-20 \\
\hline
Baseline& Softmax &38.1 &58.9  & 41.4& 31.4& 41.4&40.9\\
\hline
Raw&\multirow{7}{*}{IIF}&38.5 &59.3 &41.9 &33.5 &42.6 &41.6\\
Smooth& &\textbf{38.9} &59.6  &\textbf{42.7} &34.8 &43.7 &42.3\\
Relative& &38.6 &59.4  &42.0 &34.1 &42.8 &41.4\\
Base2&  &38.8 &\textbf{59.8}  & 42.5& 35.2 & 44.0&42.4\\
Base10&  &38.3 &59.1  & 41.7&31.5 & 41.9&41.0\\
Normit&  &38.3 &59.1  & 42.0& 31.9 & 41.9&41.3\\
Gombit&  &38.6 &59.3 & 42.1&34.5& 43.0&41.7\\
\hline
Raw&\multirow{7}{*}{IOF}&\textbf{38.9} &\textbf{59.8} &42.5&35.1&43.8&42.5\\
Smooth& &\textbf{38.9} &59.5 &42.4&\textbf{36.2} &\textbf{44.5} &\textbf{42.7}\\
Relative& &38.6 &59.5 &42.0 &33.2 &42.5 &41.6\\
Base2&  &38.6&59.4  &41.9& 33.2 & 42.0&41.3\\
Base10&  &38.5 &59.4 & 41.9& 34.9 & 43.1&41.5\\
Normit&  &38.4 &59.2  & 42.0& 32.2& 41.8&41.2\\
Gombit&  &38.5&59.3  & 41.8&32.5& 42.3&41.5\\
\hline
\end{tabular}
\label{tab:coco_abl_obj}
\end{table}

\begin{figure*}[t]
    \centering
    \includegraphics[scale=0.34]{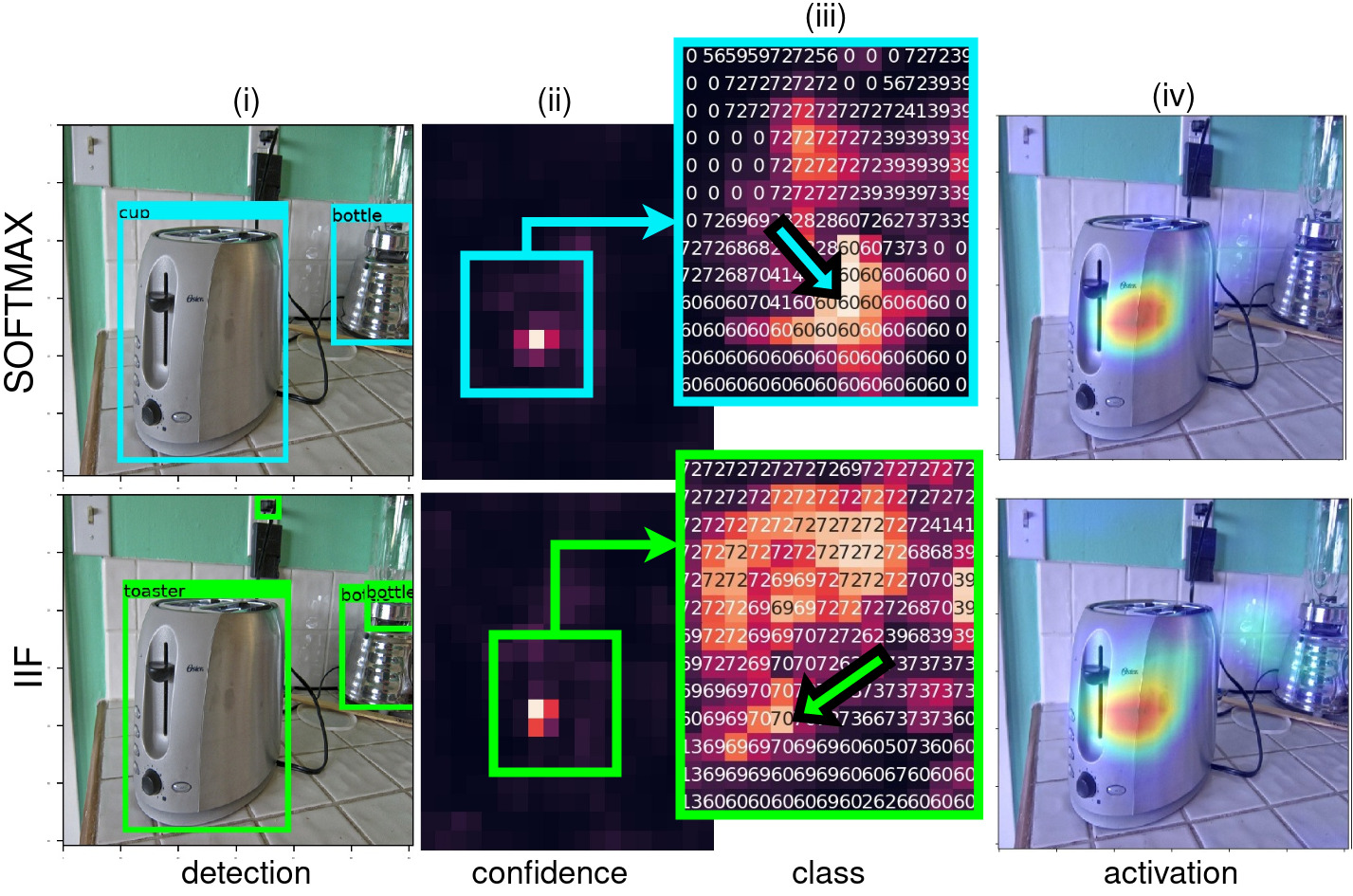}
    \caption{Softmax vs $IIF$ using YOLOV3. In (i) $IIF$ can correctly detect and classify the rare class \textit{toaster} in contrast to Softmax that predicted \textit{cup}. In (ii) the confidence heatmap is illustrated, both Softmax and $IIF$ correctly identify the center of the object. In (iii) the class heatmap for the high confidence region is shown, using the 80 classes of COCO. Softmax, predicts the class 60 with high score which is the class \textit{cup}, while $IIF$ predicts the class 70 with high score which is the correct \textit{toaster} class. In (iv) the activations are illustrated using GradCam, both Softmax and $IIF$ show large activations, but only $IIF$ has made a correct prediction. This highlights that $IIF$ performance boost is due to rare category classification rather than other factors such as localisation or confidence prediction. }
    \label{fig:yolov3_toaster}
\end{figure*}

\begin{table*}
\centering
\caption{Comparative results for Mask-RCNN, Faster-RCNN, YOLOv3 and SSD in terms of Average Precision (AP). Tail-$k$ refers to tail categories, grouped at $k$}
\begin{tabular}{cccccccc}
\hline
Model    & Method & $AP$ & $AP_{50}$ & $AP_{75}$ & tail-5 & tail-10 & tail-20\\
\hline
FasterRCNN & Softmax&37.0 &58.2  &39.9 &23.1&29.3  &35.9\\
FasterRCNN &$IIF$ &\textbf{37.5} &\textbf{59.3} &\textbf{40.5} &\textbf{25.0} &\textbf{30.3}&\textbf{36.2}\\
\hline
MaskRCNN& Softmax &38.1 &58.9  & 41.4& 31.4& 41.4&40.9\\
MaskRCNN&$IIF$ &\textbf{38.9} &\textbf{59.6}  &\textbf{42.7} &\textbf{34.8} &\textbf{43.7} &\textbf{42.3}\\
\hline
YOLOv3 &Softmax  &33.9&58.6 &35.2 &19.3 &24.8 &31.3\\
YOLOv3 & $IIF$ &\textbf{34.6} &\textbf{59.7} &\textbf{35.7} &\textbf{21.8} &\textbf{26.8}  &\textbf{32.7}\\
\hline
SSD & Softmax &25.0 & 41.5 & 25.9 & 14.7 & 17.2&22.3\\
SSD &$IIF$ &\textbf{25.7} & \textbf{43.6} &\textbf{26.4} &\textbf{18.5} &\textbf{20.0}& \textbf{24.3}\\
\hline
\end{tabular}\label{tab:resultsapar}
\end{table*}

\subsection{IIF Variants}
We expand the analysis of $IIF$ variants using MaskRCNN on the MS-COCO dataset. As seen in Table \ref{tab:coco_abl_obj}, the best variants are smooth $IIF$, raw $IOF$ and smooth $IOF$ as they achieve the best overall performance, boosting $AP$ by $0.8\%$. 
Smooth $IOF$ achieves the best performance on rare classes as it increases tail-5 by $1.1\%$, and tail-10 by $0.7\%$. In general, all $IIF$ variants boost the performance consistently, in the end, we choose to use the smooth $IIF$ variant because it generalises well in many object detection architectures and has the best performance.

\subsection{Results}
Using the smooth $IIF$, we conduct experiments with common object detectors. As the results indicate in Table \ref{tab:resultsapar}, the models equipped with our proposed $IIF$ outperform the vanilla object detectors consistently on the COCO dataset for all the object detectors.

Regarding overall $AP$, $IIF$ improves FasterRCNN by $0.5\%$, MaskRCNN by $0.8\%$, YOLOv3 and SSD by $0.7\%$. For the detection performance of $AP_{50}$, an improvement of $1.1\%$ was achieved for FasterRCNN and YOLOv3, $0.7\%$ for MaskRCNN and $2.1\%$ for SSD. Finally, regarding $AP_{75}$, $IIF$ increments performance of FasterRCNN by $0.6\%$, MaskRCNN by $1.3\%$, YOLOv3 and SSD by $0.5\%$.
The increase in performance is smaller in this task than the long-tailed instance segmentation task because the COCO dataset is less imbalanced than LVIS dataset and it contains more diverse samples per category.

Nevertheless, our method significantly increases the performance of rare categories showing consistent improvements among all the detectors. In particular, $IIF$ improves the tail-5 classes of the dataset by $1.9\%$ for FasterRCNN, by $3.4\%$ for MaskRCNN, by $2.5\%$ for YOLOv3 and by $3.8\%$ for SSD. Regarding the tail-10, our method improves FasterRCNN by $1.0\%$, MaskRCNN by $2.3\%$, YOLOv3 by $2.0\%$ and SSD by $2.8\%$. Finally, $IIF$ improves the tail-20 performance by $0.3\%$ for FasterRCNN, by $1.4\%$ for MaskRCNN, by $1.4\%$ for YOLOv3 and by $2.0\%$ for SSD.

\subsection{Model Analysis}
We use YOLOv3 \cite{redmon2018yolov3} to analyse the performance of $IIF$. YOLOv3 is a one-stage object detector that disentangles background from foreground using an objectness branch and classification branch. In Figure \ref{fig:yolov3_toaster} we show one image containing the rare class \textit{toaster} from COCO validation set. Softmax and $IIF$ both localise the object correctly as shown in (i) and have high confidence for the object's location as shown in (ii). However, only $IIF$ predicts the correct class that is \textit{toaster}, which is the class 70 in COCO, as shown (iii). Softmax on the other hand, mis-predicts the class 60, which is the \textit{cup} class with a high score. Finally, both Softmax and $IIF$ produce large activations using GradCam\cite{selvaraju2017grad} as displayed in (iv). This demonstrates that $IIF$ increases the classification ability of the network particularly for the rare classes, while it keeps its localisation and confidence prediction skill intact.

\end{document}